\providecommand{\anna}[1]{{#1}}%
\providecommand{\final}[1]{{#1}}%

\documentclass[manuscript,screen]{acmart}
\AtBeginDocument{%
  \providecommand\BibTeX{{%
    \normalfont B\kern-0.5em{\scshape i\kern-0.25em b}\kern-0.8em\TeX}}}

\setcopyright{acmcopyright}
\copyrightyear{2022}
\acmYear{2022}
\acmDOI{10.1145/3560260}

\usepackage{wrapfig}
\usepackage{tikz}
\usetikzlibrary{positioning,trees,arrows,mindmap,shapes,overlay-beamer-styles}
\usepackage{tabulary}
\usepackage{tikz-qtree}
\usepackage{multicol,comment}
\usepackage[framemethod=TikZ]{mdframed}
\tikzset{level 1 concept/.append style={font=\tiny, sibling angle=45,level distance = 75mm}}
\tikzset{level 2 concept/.append style={font=\tiny, sibling angle=45,level distance = 20mm}}

\newenvironment{itemize*}%
  {\begin{itemize}%
    \setlength{\itemsep}{.2pt}%
    \setlength{\parskip}{.2pt}%
    \setlength{\topsep}{.5pt}}%
  {\end{itemize}}

\definecolor{myOrange}{HTML}{f5a623} 
\definecolor{myPink}{HTML}{ffb2b2} 
\newcommand{\boldtext}[1]{\color{myOrange}{\textbf{#1}}~\normalcolor}
\tikzset{My Arrow Style/.style={single arrow, fill=myOrange, anchor=base, align=center,text width=2.8cm}}
\newcommand{\MyArrow}[2][]{\tikz[baseline] \node [My Arrow Style,#1] {#2};}

\usepackage{tabularx,booktabs}
\usepackage{multirow}
\usepackage[nameinlink]{cleveref}

\crefformat{section}{\S#2#1#3} %
\crefformat{subsection}{\S#2#1#3}
\crefformat{subsubsection}{\S#2#1#3}

\begin{document}

\title{QA Dataset Explosion: A Taxonomy of NLP Resources for Question Answering and Reading Comprehension} 
\author{Anna Rogers}
\email{arogers@sodas.ku.dk}
\orcid{1234-5678-9012}
\affiliation{%
  \institution{University of Copenhagen (Denmark)}
  \country{RIKEN (Japan)}
}

\author{Matt Gardner}
\email{mattgardner@microsoft.com}
\affiliation{%
  \institution{Microsoft Semantic Machines}
  \country{USA}
}

\author{Isabelle Augenstein}
\email{augenstein@di.ku.dk}
\affiliation{%
  \institution{University of Copenhagen}
  \country{Denmark}
}

\renewcommand{\shortauthors}{Rogers, Gardner and Augenstein}

\begin{abstract}

Alongside huge volumes of research on deep learning models in NLP in the recent years, there has been also much work on benchmark datasets needed to track modeling progress. Question answering and reading comprehension have been particularly prolific in this regard, with over 80 new datasets appearing in the past two years. This study is the largest survey of the field to date. We provide an overview of the various formats and domains of the current resources, highlighting the current lacunae for future work. We further discuss the current classifications of ``skills'' that question answering/reading comprehension systems are supposed to acquire, and propose a new taxonomy. The supplementary materials survey the current multilingual resources and monolingual resources for languages other than English, and we discuss the implications of over-focusing on English. The study is aimed at both practitioners looking for pointers to the wealth of existing data, and at researchers working on new resources. 
\end{abstract}


\maketitle

\section{Introduction: the Dataset Explosion}

The rapid development of NLP data in the past years can be compared to the Cambrian explosion: the time when the fossil record shows a vast increase in the number of living species. %
In the case of NLP in 2013-2020, the key ``resource'' that made this explosion possible was the widespread use of crowdsourcing, essential for the new data-hungry deep learning models. The evolutionary forces behind the explosion were (a) a desire to push more away from linguistic structure prediction and towards a (still vague) notion of ``natural language understanding'' (NLU), which different research groups pursued in different directions, and (b) the increasing practical utility of commercial NLP systems incorporating questions answering technology (for search, chatbots, personal assistants, and other applications). 
A key factor in this process is that it was a breadth-first search: there was little coordination between groups, besides keeping track of concurrent work by competitor labs. %

The result is a potpourri of datasets that is difficult to reduce to a single taxonomy, and for which it would be hard to come up with a single defining feature that would apply to all the resources. For instance, while we typically associate ``question answering'' (QA) and ``reading comprehension'' (RC) with a setting where there is an explicit question that the model is supposed to answer, even that is not necessarily the case. Some such datasets are in fact based on statements rather than questions (as in many cloze formatted datasets, see \cref{sec:format-cloze}), or on a mixture of statements and questions. 

The chief contribution of this work is a systematic review of the existing resources with respect to a set of criteria, which also broadly correspond to research questions NLP has focused on so far. After discussing the distinction between probing and information-seeking questions (\cref{sec:natural}), and the issue of question answering as a task vs format (\cref{sec:format-task}), we outline the key dimensions for the format of the existing resources: questions  (questions vs statements, \cref{sec:question-format}), answers (extractive, multi-choice, categorical and freeform, \cref{sec:answer-format}), and input evidence (in terms of its modality and amount of information \cref{sec:text-format}). \anna{Then we consider the conversational features of the current QA/RC resources (\cref{sec:discourse})}, their domain coverage (\cref{sec:domains}), and languages for which resources are available (\cref{sec:languages}). We conclude with an overview of the ``skills'' targeted by the current benchmarks (\cref{sec:reasoning}), providing an overview of the current classifications and proposing our own taxonomy (along the dimensions of inference, information retrieval, world modeling, input interpretation, and multi-step reasoning). We conclude with the discussion of the issue of ``requisite'' skills and the gaps in the current research (\cref{sec:discussion}).

For each of these criteria, we discuss how it is conceptualized in the field, with representative examples of English\footnote{Most QA/RC resources are currently in English, so the examples we cite are in English, unless specified otherwise. \Cref{sec:languages} discusses the languages represented in the current monolingual and multilingual resources, including the tendencies \& incentives for their creation.} resources of each type. What this set of criteria allows us to do is to place QA/RC work in the broader context of work on machine reasoning and linguistic features of NLP data, in a way that allows for easy connections to other approaches to NLU such as inference and entailment. It also allows us to map out the field in a way that highlights the cross-field connections (especially multi-modal NLP and commonsense reasoning) and gaps for future work to fill. %

This survey focuses exclusively on the typology of the existing resources, and its length is proof that data work on RC/QA has reached the volume at which it is no longer possible to survey in conjunction with modeling work. We refer the reader to the existing surveys~\cite{QiuChenEtAl_2019_Survey_on_Neural_Machine_Reading_Comprehension,ZhuLeiEtAl_2021_Retrieving_and_Reading_Comprehensive_Survey_on_Open-domain_Question_Answering} and tutorials~\cite{ChenYih_2020_Open-Domain_Question_Answering,RuderAvirup_2021_Multi-Domain_Multilingual_Question_Answering} for the current approaches to modeling in this area. \anna{It is also wider in scope than the existing surveys of QA data which focus on the ``skills'' they cover \cite{SugawaraAizawa_2016_Analysis_of_Prerequisite_Skills_for_Reading_Comprehension,SchlegelValentinoEtAl_2020_Framework_for_Evaluation_of_Machine_Reading_Comprehension_Gold_Standards,DzendzikVogelEtAl_2021_English_Machine_Reading_Comprehension_Datasets_Survey}, or on providing relatively detailed descriptions of datasets together with associated leaderboards and leading systems \cite{CambazogluSandersonEtAl_2020_Review_of_Public_Datasets_in_Question_Answering_Research}.}

\section{Information-seeking vs probing questions}
\label{sec:natural}

The most fundamental distinction in QA datasets is based on the communicative intent of the author of the question: was the person seeking information they did not have, or trying to test the knowledge of another person or machine?\footnote{There are many other possible communicative intents of questions in natural language, such as expressing surprise, emphasis, or sarcasm.  These do not as yet have widely-used NLP datasets constructed around them, so we do not focus on them in this survey.}  There are resources constructed from questions which appeared ``in the wild'' as a result of humans seeking information, %
while others consist of questions written by people who already knew the correct answer, for the purpose of probing NLP systems. These two kinds of questions broadly correlate with the ``tasks'' of QA and RC: QA is more often associated with information-seeking questions and RC with probing questions, and many of the other dimensions discussed in this survey tend to cluster based on this distinction. \final{The researchers working with information-seeking vs probing questions also tend to have fundamentally different motivation and research programs \cite{RodriguezBoyd-Graber_2021_Evaluation_Paradigms_in_Question_Answering}}.

Information-seeking questions tend to be written by users of some product, such as Google Search~\cite{KwiatkowskiPalomakiEtAl_2019_Natural_Questions_Benchmark_for_Question_Answering_Research,ClarkLeeEtAl_2019_BoolQ_Exploring_Surprising_Difficulty_of_Natural_YesNo_Questions}, Reddit~\cite{FanJerniteEtAl_2019_ELI5_Long_Form_Question_Answering} or community question answering sites like StackOverflow~\cite[e.g.][]{CamposOtegiEtAl_2020_DoQA-Accessing_Domain-Specific_FAQs_via_Conversational_QA} and Yahoo Answers~\cite[e.g.][]{HashemiAliannejadiEtAl_2019_ANTIQUE_Non-Factoid_Question_Answering_Benchmark} (although in some cases crowd workers are induced to write information-seeking questions~\cite{ClarkChoiEtAl_2020_TyDi_QA_Benchmark_for_Information-Seeking_Question_Answering_in_Typologically_Diverse_Languages,dasigi-etal-2021-dataset,FergusonGardnerEtAl_2020_IIRC_Dataset_of_Incomplete_Information_Reading_Comprehension_Questions}).  Most often, these questions assume no given context (\S\ref{sec:knowledge}) and are almost never posed as multiple choice (\S\ref{sec:answer-format}). Industrial research tends to focus on this category of questions, as research progress directly translates to improved products.  An appealing aspect of these kinds of questions is that they typically arise from real-world use cases, and so can be sampled to create a ``natural'' distribution of questions that people ask -- this is why a dataset created from queries issued to Google Search was called ``Natural Questions''~\cite{KwiatkowskiPalomakiEtAl_2019_Natural_Questions_Benchmark_for_Question_Answering_Research}.  However, care must be taken in saying that there exists a ``natural distribution'' over all questions: the distribution of Google Search queries is in no way representative of all questions a person typically asks in a day, and it is not clear that such a concept is even useful, as questions have a wide variety of communicative intents.

Probing questions, on the other hand, tend to be written either by exam writers (\S\ref{sec:domains}) or crowd workers~\cite[e.g.][]{RajpurkarZhangEtAl_2016_SQuAD_100000+_Questions_for_Machine_Comprehension_of_Text,RichardsonBurgesEtAl_2013_MCTest_A_Challenge_Dataset_for_the_Open-Domain_Machine_Comprehension_of_Text,DuaWangEtAl_2019_DROP_Reading_Comprehension_Benchmark_Requiring_Discrete_Reasoning_Over_Paragraphs}. 
The questions are most often written with the intent to probe understanding of a specific context, such as a paragraph or an image (\S\ref{sec:knowledge}); if a person were presented this context and wanted to extract some information from it, they would just examine the context instead of posing a question to a system. One could argue that questions written for testing \textit{human} reading comprehension constitute a ``natural'' distribution of probing questions, but this distribution is likely not ideal for testing \textit{models} (see \cref{sec:format-multi-choice}), especially if a large training set is given which can be mined for subtle spurious patterns \cite{GardnerMerrillEtAl_2021_Competency_Problems_On_Finding_and_Removing_Artifacts_in_Language_Data}.  Instead, researchers craft classes of questions that probe particular aspects of reading comprehension in machines, and typically employ crowd workers to write large collections of these questions.

These two classes of questions also tend to differ in the kinds of reasoning they require (\S\ref{sec:reasoning}).  Information-seeking questions are often ill-specified, full of ``ambiguity and presupposition''~\cite{KwiatkowskiPalomakiEtAl_2019_Natural_Questions_Benchmark_for_Question_Answering_Research}, and so real-world QA applications would arguably need to show that they can handle this kind of data. %
But while the presence of ambiguous questions or questions with presuppositions make such data more ``natural'', it also makes such data problematic as a benchmark~\cite{Boyd-Graber_2019_What_Question_Answering_can_Learn_from_Trivia_Nerds}: nearly half of the Natural Questions are estimated to be ambiguous~\cite{MinMichaelEtAl_2020_AmbigQA_Answering_Ambiguous_Open-domain_Questions}, and there are new resources specifically targeting the ambiguity challenge \cite[e.g.][]{MinMichaelEtAl_2020_AmbigQA_Answering_Ambiguous_Open-domain_Questions,ZhangChoi_2021_SituatedQA_Incorporating_Extra-Linguistic_Contexts_into_QA,SunCohenEtAl_2022_ConditionalQA_Complex_Reading_Comprehension_Dataset_with_Conditional_Answers,ChenGudipatiEtAl_2021_Evaluating_Entity_Disambiguation_and_Role_of_Popularity_in_Retrieval-Based_NLP}. Furthermore, especially when collected from search queries, information-seeking questions also tend to involve less complex reasoning than is seen in some probing datasets, as users do not expect search engines to be able to handle complex questions and so they do not ask them.  This is not to say that there are no complex questions in search-based data, but they are less frequent, while probing datasets can be specifically constructed to target one piece of the long tail in a more ``natural'' distribution.

Lastly, while we distinguish between information-seeking and probing questions, the lines are often blurry.  For example, the question \textit{``Which program at Notre Dame offers a Master of Education degree?''} could be asked by a college applicant seeking information, but it also occurs in SQuAD, a probing dataset~\cite{RajpurkarZhangEtAl_2016_SQuAD_100000+_Questions_for_Machine_Comprehension_of_Text}. \anna{The questions from Natural Questions were originally information-seeking, but when they were provided as seed questions to annotators in TopiOCQA~\cite{AdlakhaDhuliawalaEtAl_2022_TopiOCQA_Open-domain_Conversational_Question_Answering_with_Topic_Switching}, the real need for information was not present anymore in either the seed or the follow-up questions}.  When paired with single documents that likely contain the answer to the question, information-seeking datasets become much more like probing datasets~\cite{ClarkChoiEtAl_2020_TyDi_QA_Benchmark_for_Information-Seeking_Question_Answering_in_Typologically_Diverse_Languages}.  Some datasets intentionally combine elements of both, probing an initial context while at the same time eliciting information seeking questions that need additional context to be answered~\cite{FergusonGardnerEtAl_2020_IIRC_Dataset_of_Incomplete_Information_Reading_Comprehension_Questions,dasigi-etal-2021-dataset}.
\section{Task versus format}
\label{sec:format-task}
Strictly speaking, almost any NLP task can be formulated as question answering, and this is already being leveraged for model reuse and multi-task learning \cite[e.g.][]{McCannKeskarEtAl_2018_Natural_Language_Decathlon_Multitask_Learning_as_Question_Answering,weissenborn-etal-2018-jack} and zero-shot learning \cite[e.g.][]{LevySeoEtAl_2017_Zero-Shot_Relation_Extraction_via_Reading_Comprehension,abdou-etal-2019-x}. For example, machine translation could be recast as answering questions like ``What is the translation of X into German?'', and sentiment analysis -- as ``What is the sentiment of X?''. Under this view, a survey of QA datasets would encompass all NLP datasets. \anna{In such cases, QA is not a task but a format: ``a way of posing a particular problem to a machine, just as classification or natural language inference are formats''}~\cite{GardnerBerantEtAl_2019_Question_Answering_is_Format_When_is_it_Useful}.

The key distinction to keep in mind is ``how easy would it be to replace the questions in a dataset with content-free identifiers?''~\cite{GardnerBerantEtAl_2019_Question_Answering_is_Format_When_is_it_Useful}. An illustration of this heuristic is shown in Figure~\ref{fig:task-format}. Sentiment analysis is a classification task, so the questions correspond to a few labels and could easily be replaced. An NLP system does not actually need to ``understand'' the wording of the recast question, beyond the part that needs to be classified.
This heuristic is not a strict criterion, however, and the boundaries are fuzzy. Some datasets that have been published and used as QA or RC datasets can be templated with a few dozen templates \cite[e.g.][]{WestonBordesEtAl_2015_Towards_AIcomplete_question_answering_A_set_of_prerequisite_toy_tasks}. Still, such datasets have enabled undeniable progress, and will likely continue to be useful. What has changed is our awareness of how the low diversity of patterns in the training data leads to the over-reliance on these patterns \cite[][among others]{JiaLiang_2017_Adversarial_Examples_for_Evaluating_Reading_Comprehension_Systems,GevaGoldbergEtAl_2019_Are_We_Modeling_Task_or_Annotator_Investigation_of_Annotator_Bias_in_Natural_Language_Understanding_Datasetsa,McCoyPavlickEtAl_2019_Right_for_Wrong_Reasons_Diagnosing_Syntactic_Heuristics_in_Natural_Language_Inference,Linzen_2020_How_Can_We_Accelerate_Progress_Towards_Human-like_Linguistic_Generalization}.%

One should also not conflate format with reasoning types (\cref{sec:reasoning}). For example, ``extractive QA'' is often discussed as if were a cohesive problem -- however, extractive QA is an output format, and datasets using this format can differ wildly in the nature of the problems they encapsulate.

\begin{figure}[t]
	\MyArrow[text width=\textwidth]{ \textit{[FORMAT]}\hfill how easily can the questions be replaced with ids? \hfill\textit{[TASK]}}
	
	\begin{multicols}{3}
		\textit{(easy)} \\ \boldtext{Classification} \\ What is thear sentiment of <STATEMENT>? \columnbreak
		
		\textit{(doable)} \\ \boldtext{Template-filling} \\ When was <PERSON> born? \columnbreak
		
		\textit{(difficult)} \\ \boldtext{Open-ended} \\ (too many templates and/or variables)
	\end{multicols}
	\caption{When is question answering a task, and when is it a format?}
	\label{fig:task-format}
\end{figure}

\section{Format}
\label{sec:format}

This section %
describes existing datasets along the dimension of formats for \textit{questions} (the text used to query the system, \cref{sec:question-format}), \textit{answers} (the system output, \cref{sec:answer-format}), and \textit{evidence} (the source of knowledge used to derive the system output, \cref{sec:text-format}).

\subsection{Question format}
\label{sec:question-format}

\subsubsection{Natural language questions.}

Most QA and RC datasets have ``questions'' formulated as questions that a human speaker \textit{could} ask, for either information-seeking or probing purposes (see \cref{sec:natural}). They could further be described in terms of their syntactic structure: yes/no questions (\textit{Did it rain on Monday?}) wh-questions (\textit{When did it rain?}), tag questions (It rained, didn't it?), or declarative questions (It rained?)~\cite{hedberg2004meanings}. Resources with syntactically well-formed questions as question format may come with any type of answer format described in \cref{sec:answer-format}.

\subsubsection{Queries.} 

While queries are not well-formed questions, they contain pieces of information that could be interpreted as such (e.g. \textit{tallest mountain Scotland} $\longrightarrow$ \textit{which mountain is the tallest in Scotland}?). One type of relevant resources is based on logical queries (typically for tables and knowledge bases (KB)), which may then be converted to syntactically well-formed questions with templates \cite[e.g.][]{WestonBordesEtAl_2015_Towards_AIcomplete_question_answering_A_set_of_prerequisite_toy_tasks}, which may be manually edited later \cite[e.g.][]{GuKaseEtAl_2021_Beyond_IID_Three_Levels_of_Generalization_for_Question_Answering_on_Knowledge_Bases}. On the messy side of that spectrum we have search engine queries that people do \textit{not} necessarily form as either syntactically well-formed questions or as KB queries. The current datasets with ``natural'' questions use filters to remove such queries ~\cite{BajajCamposEtAl_2016_MS_MARCO_Human_Generated_MAchine_Reading_COmprehension_Dataset,KwiatkowskiPalomakiEtAl_2019_Natural_Questions_Benchmark_for_Question_Answering_Research}. How we could study the full range of human interactions with search engines is an open problem at the boundary of QA and IR, but there is at least one attempt to provide a resource of search engine queries annotated for well-formedness~\cite{FaruquiDas_2018_Identifying_Well-formed_Natural_Language_Questions}.

\begin{table}[t]
	\caption{\anna{Question formats of question answering and reading comprehension datasets}}
	\label{tab:formats-questions}
	\centering
	\begin{tabular}{p{1.8cm} p{1.5cm} p{3.3cm} p{1cm} p{4.22cm}}
		\toprule    
		Evidence & Format & Question & Answer & Example datasets \\
		\midrule    
		\multirow{5}{2cm}{Einstein was born in 1879.} & Questions & When was Einstein born? & 1879 & SQuAD~\cite{RajpurkarZhangEtAl_2016_SQuAD_100000+_Questions_for_Machine_Comprehension_of_Text}, RACE~\cite{LaiXieEtAl_2017_RACE_Large-scale_ReAding_Comprehension_Dataset_From_Examinations} \\
		& Queries & Which year Einstein born & 1879 
		& %
		generated queries in BEIR~\cite{ThakurReimersEtAl_2021_BEIR_Heterogeneous_Benchmark_for_Zero-shot_Evaluation_of_Information_Retrieval_Models} \\          
		& Cloze & Einstein was born in \underline{\hspace{.5cm}}. & 1979 & CNN/Daily Mail~\cite{HermannKociskyEtAl_2015_Teaching_Machines_to_Read_and_Comprehend}, CBT~\cite{HillBordesEtAl_2015_Goldilocks_Principle_Reading_Childrens_Books_with_Explicit_Memory_Representations} \\          
		& Completion & Einstein was born ... & in 1879  & SWAG~\cite{ZellersBiskEtAl_2018_SWAG_A_Large-Scale_Adversarial_Dataset_for_Grounded_Commonsense_Inference}, RocStories~\cite{MostafazadehRothEtAl_2017_LSDSem_2017_Shared_Task_The_Story_Cloze_Test} \\          
		\bottomrule
	\end{tabular}
\end{table}    

\subsubsection{Cloze format.} 
\label{sec:format-cloze} 

Cloze statements are neither questions nor queries, but simply sentences with a masked span which, similarly to extractive QA format (see \cref{sec:format-extractive}), the model needs to predict. The key difference with questions or queries is that cloze statements are simply excerpts from the evidence document (or some other related text), rather than something specifically formulated for information extraction. The sentences to be converted to Cloze ``questions'' have been identified as:

\begin{itemize}
    \item simply sentences contained within the text~\cite{LongBengioEtAl_2017_World_Knowledge_for_Reading_Comprehension_Rare_Entity_Prediction_with_Hierarchical_LSTMs_Using_External_Descriptions};
    \item designating an excerpt as the ``text", and the sentence following it as the ``question''~\cite{HillBordesEtAl_2015_Goldilocks_Principle_Reading_Childrens_Books_with_Explicit_Memory_Representations,PapernoKruszewskiEtAl_2016_LAMBADA_dataset_Word_prediction_requiring_broad_discourse_context};
    \item given a text and summary of that text, use the summary as the question~\cite{HermannKociskyEtAl_2015_Teaching_Machines_to_Read_and_Comprehend}.
\end{itemize}

The Cloze format has been often used to test the knowledge of entities (CNN/Daily Mail~\cite{HermannKociskyEtAl_2015_Teaching_Machines_to_Read_and_Comprehend}, WikiLinks Rare Entity ~\cite{LongBengioEtAl_2017_World_Knowledge_for_Reading_Comprehension_Rare_Entity_Prediction_with_Hierarchical_LSTMs_Using_External_Descriptions}). Other datasets targeted a mixture of named entities, common nouns, verbs (CBT~\cite{HillBordesEtAl_2015_Goldilocks_Principle_Reading_Childrens_Books_with_Explicit_Memory_Representations}, LAMBADA~\cite{PapernoKruszewskiEtAl_2016_LAMBADA_dataset_Word_prediction_requiring_broad_discourse_context}). While the early datasets focused on single words or entities to be masked, there are also resources masking sentences in the middle of the narrative \cite{KongGangalEtAl_2020_SCDE_Sentence_Cloze_Dataset_with_High_Quality_Distractors_From_Examinations,CuiLiuEtAl_2020_Sentence_Cloze_Dataset_for_Chinese_Machine_Reading_Comprehension}.

The Cloze format has the advantage that these datasets can be created programmatically, resulting in quick and inexpensive data collection (although it can also be expensive if additional filtering is done to ensure answerability and high question quality~\cite{PapernoKruszewskiEtAl_2016_LAMBADA_dataset_Word_prediction_requiring_broad_discourse_context}). %
But Cloze questions are not technically ``questions'', and so do not \textit{directly} target the QA task. The additional limitation is that only the relations within a given narrow context can be targeted, and it is difficult to control the kind of information that is needed to fill in the Cloze: it could simply be a collocation, or a generally-known fact -- or some unique relation only expressed within this context.

The Cloze format is currently resurging in popularity also as a way to evaluate masked language models~\cite{Ettinger_2020_What_BERT_is_not_Lessons_from_new_suite_of_psycholinguistic_diagnostics_for_language_models,Goldberg_2019_Assessing_BERTs_Syntactic_Abilities}, as fundamentally the Cloze task is what these models are doing in pre-training. 

\subsubsection{Story completion.}
\label{sec:format-ending}

A popular format in commonsense reasoning is the choice of the alternative endings for the passage (typically combined with multi-choice answer format (see \cref{sec:format-multi-choice})). It could be viewed as a variation of Cloze format, but many Cloze resources have been generated automatically from existing texts, while choice-of-ending resources tend to be crowdsourced for this specific purpose. Similarly to the Cloze format, the ``questions'' are not necessarily linguistically well-formed questions. They may be unfinished sentences (as in SWAG~\cite{ZellersBiskEtAl_2018_SWAG_A_Large-Scale_Adversarial_Dataset_for_Grounded_Commonsense_Inference} and HellaSWAG~\cite{ZellersHoltzmanEtAl_2019_HellaSwag_Can_Machine_Really_Finish_Your_Sentence}) or short texts (as in RocStories~\cite{MostafazadehRothEtAl_2017_LSDSem_2017_Shared_Task_The_Story_Cloze_Test}) to be completed. 

\subsection{Answer format}
\label{sec:answer-format}

The outputs of the current text-based datasets can be categorized as extractive (\cref{sec:format-extractive}), multi-choice (\cref{sec:format-multi-choice}), categorical (\cref{sec:format-categorical}), or freeform (\cref{sec:format-freeform}), as shown in \autoref{tab:formats}.

\begin{table}[t]
    \caption{Answer formats of question answering and reading comprehension datasets}
    \label{tab:formats}
    \centering
    \begin{tabular}{p{1.8cm} p{1.8cm} p{3.8cm} p{2.1cm} p{4cm}}
\toprule    
    Evidence & Format & Question & Answer(s) & Example datasets \\
\midrule    
\multirow{5}{2cm}{Einstein was born in 1879.} & Extractive & When was Einstein born? & 1879 (token 5) & SQuAD~\cite{RajpurkarZhangEtAl_2016_SQuAD_100000+_Questions_for_Machine_Comprehension_of_Text}, NewsQA~\cite{trischler-etal-2017-newsqa} \\
& Multi-choice & When was Einstein born? & (a) 1879, (b) 1880  
& RACE~\cite{LaiXieEtAl_2017_RACE_Large-scale_ReAding_Comprehension_Dataset_From_Examinations} \\          
& Categorical & Was Einstein born in 1880? & No & BoolQ~\cite{ClarkLeeEtAl_2019_BoolQ_Exploring_Surprising_Difficulty_of_Natural_YesNo_Questions} \\          
& Freeform & When was Einstein born? & 1879 (generated) & MS MARCO~\cite{BajajCamposEtAl_2016_MS_MARCO_Human_Generated_MAchine_Reading_COmprehension_Dataset}, CoQA~\cite{ReddyChenEtAl_2019_CoQA_Conversational_Question_Answering_Challenge} \\          
\bottomrule
    \end{tabular}
\end{table}    

\subsubsection{Extractive format.} 
\label{sec:format-extractive}
Given a source of evidence and a question, the task is to predict the part of the evidence (a span, in case of a text) which is a valid answer for the question. This format is very popular both thanks to its clear relevance for QA applications, and the relative ease of creating such data (questions need to be written or collected, but answers only need to be selected in the evidence). \final{While most extractive QA resources assume that there is a single correct answer, more recently the task of multi-span QA has been proposed \cite{ZhuAhujaEtAl_2020_Question_Answering_with_Long_Multiple-Span_Answers,LiTomkoEtAl_2022_MultiSpanQA_Dataset_for_Multi-Span_Question_Answering}.}
    
In its classic formulation extractive QA is the task behind search engines. The connection is very clear in early QA research: the stated goal of the first TREC QA competition in 2000 was ``to foster research
that would move retrieval systems closer to \textit{information} retrieval as opposed to \textit{document} retrieval''~\cite{VoorheesTice_2000_Building_question_answering_test_collection}. To answer questions like ``Where is the Taj Mahal?'' given a large collection of documents, the participating systems had to rank the provided documents and the candidate answer spans within the documents, and return the best five. Some of the more recent QA datasets also provide a collection of candidate documents rather than a single text (see \cref{sec:knowledge}), \final{and answer selection is also viewed as a separate task with its own resources \cite{NakovMarquezEtAl_2015_SemEval-2015_Task_3_Answer_Selection_in_Community_Question_Answering,GargVuEtAl_2020_TANDA_Transfer_and_Adapt_Pre-Trained_Transformer_Models_for_Answer_Sentence_Selection}.}

A step back into this direction came with the introduction of unanswerable questions~\cite{RajpurkarJiaEtAl_2018_Know_What_You_Dont_Know_Unanswerable_Questions_for_SQuAD,AcharyaJariwalaEtAl_2019_VQD_Visual_Query_Detection_In_Natural_Scenes,AsaiChoi_2021_Challenges_in_Information-Seeking_QA_Unanswerable_Questions_and_Paragraph_Retrieval}: the questions that target the same context as the regular questions, but do not have an answer in that context. With the addition of unanswerable questions, systems trained on extractive datasets can be used as a component of a search engine: first the candidate documents are assessed for whether they can be used to answer a given question, and then the span prediction is conducted on the most promising candidates. It is however possible to achieve search-engine-like behavior even without unanswerable questions~\cite{chen-etal-2017-reading,clark-gardner-2018-simple}.
    
Many extractive datasets are ``probing'' in that the questions were written by the people who already knew the answer, but, as the datasets based on search engine queries show, it does not have to be that way. %
A key advantage of the extractive format is that only the questions need to be written, and the limited range of answer options means that it is easier to define what an acceptable correct answer is. A key disadvantage is that it limits the kinds of questions that can be asked to questions with answers directly contained in the text. While it is possible to pose rather complex questions (\cref{sec:multi-step}), it is hard to use this format for any interpretation of the facts of the text, any meta-analysis of the text or its author's intentions, or inference to unstated propositions.

\subsubsection{Multi-choice format.} 
\label{sec:format-multi-choice}
Multiple choice questions are questions for which a small number of answer options are given as part of the question text itself.  Many existing multi-choice datasets are expert-written, stemming from school examinations (e.g. RACE~\cite{LaiXieEtAl_2017_RACE_Large-scale_ReAding_Comprehension_Dataset_From_Examinations}, CLEF QA~\cite{PenasUngerEtAl_2014_Overview_of_CLEF_Question_Answering_Track_2014}). This format has also been popular in resources targeting world knowledge and commonsense information (typically based on crowdsourced narratives): e.g. MCTest~\cite{RichardsonBurgesEtAl_2013_MCTest_A_Challenge_Dataset_for_the_Open-Domain_Machine_Comprehension_of_Text}, MCScript~\cite{OstermannRothEtAl_2018_SemEval-2018_Task_11_Machine_Comprehension_Using_Commonsense_Knowledge}, RocStories~\cite{MostafazadehRothEtAl_2017_LSDSem_2017_Shared_Task_The_Story_Cloze_Test}.

The advantage of this format over the extractive one is that the answers are no longer restricted to something explicitly stated in the text, which enables a much wider range of questions (including commonsense and implicit information). The question writer also has full control over the available options, and therefore over the kinds of reasoning that the test subject would need to be capable of. This is why this format has a long history in human education. Evaluation is also straightforward, unlike with freeform answers. The disadvantage is that writing good multi-choice questions is not easy, and if the incorrect options are easy to rule out -- the questions are not discriminative.\footnote{The STARC annotation scheme~\cite{BerzakMalmaudEtAl_2020_STARC_Structured_Annotations_for_Reading_Comprehension} is a recent proposal for controlling the quality of multi-choice questions by requiring that there are four answers, one of which is correct, one is based on a misunderstanding of the text span with the evidence for the correct answer, one is based on a distractor span, and one is plausible but unsupported by the evidence. This would allow studying the reasoning strategies of the models, but more studies are needed to show that we can generate these different types of incorrect answers at sufficient scale and without introducing extra spurious patterns.}

Since multi-choice questions have been extensively used in education, there are many insights into how to write such questions in a way that would best test \textit{human students}, both for low-level and high-level knowledge~\cite{Bailey_2018_Multiple-Choice_Item_Format,MarchamTurnbeaughEtAl_2018_Developing_Certification_Exam_Questions_More_Deliberate_Than_You_May_Think,BoneProsser_2020_Multiple_choice_questions_introductory_guide,MacFarlaneBoulet_2017_Multiple-Choice_Tests_Can_Support_Deep_Learning}. 
However, it is increasingly clear that humans and machines do not necessarily find the same things difficult, which complicates direct comparisons of their performance. In particular, teachers are instructed to ensure that all the answer options items are plausible, and given in the same form \cite[][p.4]{BoneProsser_2020_Multiple_choice_questions_introductory_guide}. This design could make the questions easy for a model backed with collocation information from a language model. However, NLP systems can be distracted by shallow lexical matches~\cite{JiaLiang_2017_Adversarial_Examples_for_Evaluating_Reading_Comprehension_Systems} or nonsensical adversarial inputs~\cite{WallaceFengEtAl_2019_Universal_Adversarial_Triggers_for_Attacking_and_Analyzing_NLP}, and be insensitive to at least some meaning-altering perturbations~\cite{RychalskaBasajEtAl_2018_Does_it_care_what_you_asked_Understanding_Importance_of_Verbs_in_Deep_Learning_QA_System}. For humans, such options that would be easy to reject. %

Humans may also respond differently when primed with different types of prior questions and/or when they are tired. QuAIL~\cite{RogersKovalevaEtAl_2020_Getting_Closer_to_AI_Complete_Question_Answering_Set_of_Prerequisite_Real_Tasks} made the first attempt to combine questions based on the textual evidence, world knowledge, and unanswerable questions, finding that this combination is difficult in human evaluation: if exposed to all three question types, humans struggle with making an educated guess vs marking the question as unanswerable, while models do not.

\subsubsection{Categorical format.} 
\label{sec:format-categorical}

We describe as ``categorical'' any format where the answers come from a strictly pre-defined set of options. As long as the set is limited to a semantic type with a clear similarity function (e.g. dates, numbers), we can have the benefit of automated evaluation metrics without the limitations of the extractive format. Other examples include rating questions for answerability %
~\cite{RajpurkarJiaEtAl_2018_Know_What_You_Dont_Know_Unanswerable_Questions_for_SQuAD} %
\final{and rating statements as true/false with respect to a given context \cite{CiosiciCecilEtAl_2021_Perhaps_PTLMs_Should_Go_to_School_-_Task_to_Assess_Open_Book_and_Closed_Book_QA}.}

Perhaps the most salient example of the categorical answer format is boolean questions, for which the most popular resource is currently BoolQ~\cite{ClarkLeeEtAl_2019_BoolQ_Exploring_Surprising_Difficulty_of_Natural_YesNo_Questions}. It was collected as ``natural'' information-seeking questions in Google search queries similarly to Natural Questions~\cite{KwiatkowskiPalomakiEtAl_2019_Natural_Questions_Benchmark_for_Question_Answering_Research}. Other resources not focusing on boolean questions specifically may also include them (e.g. MS MARCO~\cite{BajajCamposEtAl_2016_MS_MARCO_Human_Generated_MAchine_Reading_COmprehension_Dataset}, bAbI~\cite{WestonBordesEtAl_2015_Towards_AIcomplete_question_answering_A_set_of_prerequisite_toy_tasks}, QuAC~\cite{ChoiHeEtAl_2018_QuAC_Question_Answering_in_Context}). %

Another kind of categorical output format is when the set of answers seen during training is used as the set of allowed answers at test time. This allows for simple prediction -- final prediction is a classification problem -- but is quite limiting in that no test question can have an unseen answer.  Visual question answering datasets commonly follow this pattern (e.g. VQA \cite{AntolAgrawalEtAl_2015_VQA_Visual_Question_Answering}, GQA \cite{HudsonManning_2019_GQA_New_Dataset_for_Real-World_Visual_Reasoning_and_Compositional_Question_Answering}, CLEVR \cite{JohnsonHariharanEtAl_2017_Clevr_diagnostic_dataset_for_compositional_language_and_elementary_visual_reasoning}). 

\subsubsection{Freeform format.} 
\label{sec:format-freeform}
The most natural setting for human QA is to generate the answer independently rather than choose from the evidence or available alternatives. This format allows for asking any kinds of questions, and any other format can be instantly converted to it by having the system generate rather than select the available ``gold'' answer.

The problem is that the ``gold'' answer is probably not the only correct one, which makes evaluation difficult. Most questions have many correct or acceptable answers, and they would need to be evaluated on at least two axes: linguistic fluency and factual correctness. Both of these are far from being solved. On the factual side, it is possible to get high ROUGE-L scores on ELI5~\cite{FanJerniteEtAl_2019_ELI5_Long_Form_Question_Answering} with answers conditioned on irrelevant documents~\cite{KrishnaRoyEtAl_2021_Hurdles_to_Progress_in_Long-form_Question_Answering}, and even human experts find it hard to formulate questions so as to exactly specify the desired level of answer granularity, and to avoid presuppositions and ambiguity~\cite{Boyd-Graber_2019_What_Question_Answering_can_Learn_from_Trivia_Nerds}. On the linguistic side, evaluating generated language is a huge research problem in itself~\cite{vanderLeeGattEtAl_2019_Best_practices_for_human_evaluation_of_automatically_generated_text,CelikyilmazClarkEtAl_2020_Evaluation_of_Text_Generation_Survey}, and annotators struggle with longer answers~\cite{KrishnaRoyEtAl_2021_Hurdles_to_Progress_in_Long-form_Question_Answering}. There are also sociolinguistic considerations: humans answer the same question differently depending on the context and their background, which should not be ignored~\cite{Rogers_2021_Changing_World_by_Changing_Data}).

So far the freeform format has not been very popular. Perhaps the best-known example is MS MARCO~\cite{BajajCamposEtAl_2016_MS_MARCO_Human_Generated_MAchine_Reading_COmprehension_Dataset}, based on search engine queries with human-generated answers (written as summaries of provided Web snippets), in some cases with several answers per query. Since 2016, the dataset has grown\footnote{\url{https://microsoft.github.io/msmarco/}} to a million queries and is now accompanied with satellite IR tasks (ranking, keyword extraction). For NarrativeQA~\cite{KociskySchwarzEtAl_2018_NarrativeQA_Reading_Comprehension_Challenge}, crowd workers wrote both questions and answers based on book summaries.
CoQA~\cite{ReddyChenEtAl_2019_CoQA_Conversational_Question_Answering_Challenge} is a collection of dialogues of questions and answers from crowd workers, with additional step for answer verification and collecting multiple answer variants. The writers were allowed to see the evidence, and so the questions are not information-seeking, but the workers were dynamically alerted to avoid words directly mentioned in the text. ELI5~\cite{FanJerniteEtAl_2019_ELI5_Long_Form_Question_Answering} is a collection of user questions and long-form abstractive answers from the ``Explain like I'm 5'' subreddit, coupled with Web snippet evidence.

There is a lot of work to be done on evaluation for freeform QA. As a starting point, \citet{ChenStanovskyEtAl_2019_Evaluating_Question_Answering_Evaluation} evaluate the existing automated evaluation metrics (BLEU, ROUGE, METEOR, F1) for extractive and multi-choice questions converted to freeform format, concluding that these metrics may be used for some of the existing data, but they limit the kinds of questions that can be posed, and, since they rely on lexical matches, they necessarily do poorly for the more abstractive answers. They argue for developing new metrics based on representation similarity rather than ngram matches~\cite{ChenStanovskyEtAl_2020_MOCHA_Dataset_for_Training_and_Evaluating_Generative_Reading_Comprehension_Metrics}, although the current implementations are far from perfect. %

\vspace{2em}

To conclude the discussion of answer formats in QA/RC, let us note that, as with other dimensions for characterizing existing resources, these formats do not form a strict taxonomy based on one coherent principle. Conceptually, the task of extractive QA could be viewed as a multi-choice one: the choices are simply all the possible spans in the evidence document (although most of them would not make sense to humans). The connection is obvious when these options are limited in some way: for example, the questions in CBT~\cite{HillBordesEtAl_2015_Goldilocks_Principle_Reading_Childrens_Books_with_Explicit_Memory_Representations} are extractive (Cloze-style), but the system is provided with 10 possible entities from which to choose the correct answer, which makes also it a multi-choice dataset.

If the goal is general language ``understanding'', we arguably do not even want to impose strict format boundaries. %
To this end, UnifiedQA~\cite{KhashabiKhotEtAl_2020_UnifiedQA_Crossing_Format_Boundaries_With_Single_QA_System} proposes a single ``input'' format to which they convert extractive, freeform, categorical (boolean) and multi-choice questions from 20 datasets, showing that cross-format training often outperforms models trained solely in-format.

\subsection{Evidence format} 
\label{sec:text-format}

By ``evidence'' or ``context'', we mean whatever the system is supposed to ``understand'' or use to derive the answer from (including but not limited to texts in natural language). QA/RC resources can be characterized in terms of the modality of their input evidence (\cref{sec:media}), its amount (\cref{sec:knowledge}), and dynamic (conversational) vs static nature (\cref{sec:discourse}).

\subsubsection{Modality}
\label{sec:media}

While QA/RC is traditionally associated with natural language texts or structured knowledge bases, research has demonstrated the success of multi-modal approaches for QA %
(audio, images, and even video). Each of these areas is fast growing, and multimedia work may be key to overcoming issues with some implicit knowledge that is not ``naturally'' stated in text-based corpora~\cite{BiskHoltzmanEtAl_2020_Experience_Grounds_Language}. 

\textbf{Unstructured text.} Most resources described as RC benchmarks \cite[e.g.][]{RajpurkarZhangEtAl_2016_SQuAD_100000+_Questions_for_Machine_Comprehension_of_Text,RichardsonBurgesEtAl_2013_MCTest_A_Challenge_Dataset_for_the_Open-Domain_Machine_Comprehension_of_Text} have textual evidence in natural language, while many QA resources come with multiple excerpts as knowledge sources (e.g.~\cite{BajajCamposEtAl_2016_MS_MARCO_Human_Generated_MAchine_Reading_COmprehension_Dataset, ClarkCowheyEtAl_2018_Think_you_have_Solved_Question_Answering_Try_ARC_AI2_Reasoning_Challenge}). See \cref{sec:knowledge} for more discussions of the variation in the amount of text that is given as the context in a dataset.

\textbf{Semi-structured text.} A fast-growing area is QA based on information from tables, as in  WikiTableQuestions~\cite{PasupatLiang_2015_Compositional_Semantic_Parsing_on_Semi-Structured_Tables} and TableQA~\cite{VakulenkoSavenkov_2017_TableQA_Question_Answering_on_Tabular_Data}. At least two such resources have supporting annotations for attention supervision: SQL queries in WikiSQL~\cite{ZhongXiongEtAl_2017_Seq2SQL_Generating_Structured_Queries_from_Natural_Language_using_Reinforcement_Learning}, operand information in WikiOps~\cite{ChoAmplayoEtAl_2018_Adversarial_TableQA_Attention_Supervision_for_Question_Answering_on_Tables}. Most of them are based on Wikipedia, where the tables are relatively simple, but AIT-QA \cite{KatsisChemmengathEtAl_2022_AIT-QA_Question_Answering_Dataset_over_Complex_Tables_in_Airline_Industry} presents more complex tables from airline industry. \final{Another direction bridging text with visual modality is RC for structured text, which takes into account the layout of the document (e.g. WebSRC~ \cite{ChenZhaoEtAl_2021_WebSRC_Dataset_for_Web-Based_Structural_Reading_Comprehensiona}).}

\textbf{Structured knowledge.} Open-domain QA with a structured knowledge source is an alternative to looking for answers in text corpora, except that in this case, the model has to explicitly ``interpret'' the question by converting it to a query (e.g. by mapping the text to a triplet of entities and relation, as in WikiReading~\cite{HewlettLacosteEtAl_2016_WikiReading_Novel_Large-scale_Language_Understanding_Task_over_Wikipedia}). The questions can be composed based on the target structured information, as in SimpleQuestions~\cite{BordesUsunierEtAl_2015_Large-scale_Simple_Question_Answering_with_Memory_Networks} or Event-QA~\cite{CostaGottschalkEtAl_2020_Event-QA_Dataset_for_Event-Centric_Question_Answering_over_Knowledge_Graphs}. The process is reversed in FreebaseQA~\cite{JiangWuEtAl_2019_FreebaseQA_New_Factoid_QA_Data_Set_Matching_Trivia-Style_Question-Answer_Pairs_with_Freebase}, which collects independently authored Trivia questions and filters them to identify the subset that can be answered with Freebase information. The datasets may target a specific knowledge base: a general one such as WikiData~\cite{HewlettLacosteEtAl_2016_WikiReading_Novel_Large-scale_Language_Understanding_Task_over_Wikipedia} or Freebase~\cite{berant-etal-2013-semantic,JiangWuEtAl_2019_FreebaseQA_New_Factoid_QA_Data_Set_Matching_Trivia-Style_Question-Answer_Pairs_with_Freebase}, or one restricted to a specific application domain~\cite{hemphill-etal-1990-atis,wong-mooney-2006-learning}. 
    
\textbf{Images.} While much of this work is presented in the computer vision community, the task of multi-modal QA (combining visual and text-based information) is a challenge for both computer vision and NLP communities. The complexity of the verbal component is on a sliding scale: from simple object labeling, as in MS COCO~\cite{LinMaireEtAl_2014_Microsoft_COCO_Common_Objects_in_Context} to complex compositional questions, as in GQA~\cite{HudsonManning_2019_GQA_New_Dataset_for_Real-World_Visual_Reasoning_and_Compositional_Question_Answering} and ChartQA \cite{MasryLongEtAl_2022_ChartQA_Benchmark_for_Question_Answering_about_Charts_with_Visual_and_Logical_Reasoning}. 

While the NLP community is debating the merits of the ``natural'' information-seeking vs probing questions and both types of data are prevalent, (see \cref{sec:natural}), for visual QA the situation is skewed towards the probing questions, since most of them are based on large image bases such as COCO, Flickr or ImageNet which do not come with any independently occurring text. Accordingly, the verbal part may be created by crowdworkers based on the provided images (e.g.~\cite{SuhrLewisEtAl_2017_Corpus_of_Natural_Language_for_Visual_Reasoning}), or (more frequently) generated, e.g. AQUA~\cite{GarciaYeEtAl_2020_Dataset_and_Baselines_for_Visual_Question_Answering_on_Art}, IQA~\cite{GordonKembhaviEtAl_2018_Iqa_Visual_question_answering_in_interactive_environments}. In VQG-Apple~\cite{PatelBindalEtAl_2020_Generating_Natural_Questions_from_Images_for_Multimodal_Assistants} the crowd workers were provided with an image and asked to write questions one \textit{might} ask a digital assistant about that image, but the paper does not provide analysis of how realistic the result is. 

\textbf{Audio.} ``Visual QA'' means ``answering questions \textit{about} images. Similarly, there is a task for QA \textit{about} audio clips. DAQA~\cite{FayekJohnson_2020_Temporal_Reasoning_via_Audio_Question_Answering} is a dataset consisting of audio clips and questions about what sounds can be heard in the audio, and in what order. As with most VQA work, the questions are synthetic.

Interestingly, despite the boom of voice-controlled digital assistants that answer users' questions (such as Siri or Alexa), public data for purely audio-based question answering is so far a rarity: the companies developing such systems undoubtedly have a lot of customer data, but releasing portions of it would be both ethically challenging and not aligned with their business interests. The result is that in audio QA the QA part seems to be viewed as a separate, purely text-based component of a pipeline with speech-to-text input and text-to-speech output. That may not be ideal, because in real conversations, humans take into account prosodic cues for disambiguation, but so far, there are few such datasets, making this a promising future research area. So far there are two small-scale datasets produced by human speakers: one based on TOEFL listening comprehension data~\cite{TsengShenEtAl_2016_Towards_Machine_Comprehension_of_Spoken_Content_Initial_TOEFL_Listening_Comprehension_Test_by_Machine}, and one for a Chinese SquAD-like dataset~\cite{LeeWangEtAl_2018_ODSQA_Open-Domain_Spoken_Question_Answering_Dataset}. Spoken-SQuAD~\cite{LiWuEtAl_2018_Spoken_SQuAD_Study_of_Mitigating_Impact_of_Speech_Recognition_Errors_on_Listening_Comprehension} and Spoken-CoQA~\cite{YouChenEtAl_2020_Towards_Data_Distillation_for_End-to-end_Spoken_Conversational_Question_Answering} have audio clips generated with a text-to-speech engine.

Another challenge for audio-based QA is the conversational aspect: questions may be formulated differently depending on previous dialogue. See \cref{sec:discourse} for an overview of the text-based work in that area.

\textbf{Video.} QA on videos is also a growing research area. Existing datasets are based on movies (MovieQA~\cite{TapaswiZhuEtAl_2016_MovieQA_Understanding_Stories_in_Movies_through_Question-Answering}, MovieFIB~\cite{MaharajBallasEtAl_2017_dataset_and_exploration_of_models_for_understanding_video_data_through_fill-in-the-blank_question-answering}), TV shows (TVQA~\cite{LeiYuEtAl_2018_TVQA_Localized_Compositional_Video_Question_Answering}), games (MarioQA~\cite{MunSeoEtAl_2017_MarioQA_Answering_Questions_by_Watching_Gameplay_Videos}), cartoons (PororoQA~\cite{KimHeoEtAl_2017_DeepStory_Video_Story_QA_by_Deep_Embedded_Memory_Networks}), and tutorials (TurorialVQA~\cite{ColasKimEtAl_2020_TutorialVQA_Question_Answering_Dataset_for_Tutorial_Videos}). Some are ``multi-domain'': VideoQA~\cite{ZhuXuEtAl_2017_Uncovering_Temporal_Context_for_Video_Question_Answering} comprises clips from movies, YouTube videos and cooking videos, while TGIF-QA is based on miscellaneous GIFs~\cite{JangSongEtAl_2017_TGIF-QA_Toward_Spatio-Temporal_Reasoning_in_Visual_Question_Answering}.

As with other multimedia datasets, the questions in video QA datasets are most often generated \cite[e.g.][]{ZhuXuEtAl_2017_Uncovering_Temporal_Context_for_Video_Question_Answering,JangSongEtAl_2017_TGIF-QA_Toward_Spatio-Temporal_Reasoning_in_Visual_Question_Answering} and the source of text used for generating those question matters a lot: the audio descriptions tend to focus on visual features, and text summaries focus on the plot~\cite{LeiYuEtAl_2018_TVQA_Localized_Compositional_Video_Question_Answering}. TVQA questions are written by crowd workers, but they are still clearly probing rather than information-seeking. It is an open problem what a ``natural'' video QA would even be like: questions asked by someone who is deciding whether to watch a video? Questions asked to replace watching a video? Questions asked by movie critics?

\textbf{Other combinations.} While most current datasets fall into one of the above groups, there are also other combinations. For instance, HybridQA~\cite{ChenZhaEtAl_2020_HybridQA_Dataset_of_Multi-Hop_Question_Answering_over_Tabular_and_Textual_Data} target the information combined from text and tables, and MultiModalQA~\cite{TalmorYoranEtAl_2021_MultimodalQA_Complex_Question_Answering_Over_Text_Tables_and_Images} adds images to that setting. MovieQA~\cite{TapaswiZhuEtAl_2016_MovieQA_Understanding_Stories_in_Movies_through_Question-Answering} has different ``settings'' based on what combination of input data is used (plots, subtitles, video clips, scripts, and DVS transcription).

The biggest challenge for all multimodal QA work is to ensure that all the input modalities are actually necessary to answer the question~\cite{ThomasonGordonEtAl_2019_Shifting_Baseline_Single_Modality_Performance_on_Visual_Navigation_QA}: it may be possible to pick the most likely answer based only on linguistic features, or detect the most salient object in an image while ignoring the question. 
After that, there is the problem of ensuring that \textit{all} that multimodal information needs to be taken into account: for instance, if a model learns to answer questions about presence of objects based on a single image frame instead of the full video, it may answer questions incorrectly when the object is added/removed during the video. See also \cref{sec:discussion-required-skills} for discussion of the problem of ``required'' skills.

\subsubsection{Amount of evidence}
\label{sec:knowledge}

The second dimension for characterizing the input of a QA/RC dataset is how much evidence the system is provided with. Here, we observe the following options: 

\begin{figure}
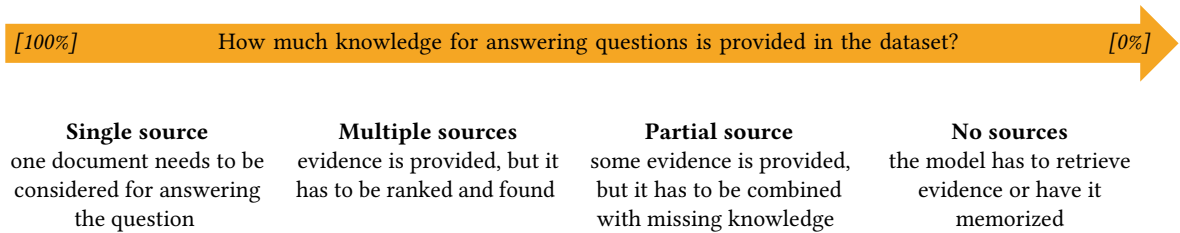


\MyArrow[text width=\textwidth]{ \textit{[100\%]}\hfill How much knowledge for answering questions is provided in the dataset? \hfill\textit{[0\%]}}
\begin{multicols}{4}
\textbf{Single source} \\ one document needs to be considered for answering the question \columnbreak \\
\textbf{Multiple sources} \\ evidence is provided, but it has to be ranked and found \columnbreak \\
\textbf{Partial source} \\ some evidence is provided, but it has to be combined with missing knowledge \columnbreak \\
\textbf{No sources} \\ the model has to retrieve evidence or have it memorized
\end{multicols}
\caption{Sources of knowledge for answering the questions.}
\label{fig:knowledge-sources}
\end{figure}

\begin{itemize*}
    \item \textbf{Single source:} the model needs to consider a pre-defined tuple of a document and a question (and, depending on the format, answer option(s)). Most RC datasets such as RACE~\cite{LaiXieEtAl_2017_RACE_Large-scale_ReAding_Comprehension_Dataset_From_Examinations} and SQuAD~\cite{RajpurkarZhangEtAl_2016_SQuAD_100000+_Questions_for_Machine_Comprehension_of_Text} fall in this category. %
    A version of this are resources with a \textit{long} input text, such as complete books~\cite{KociskySchwarzEtAl_2018_NarrativeQA_Reading_Comprehension_Challenge} or academic papers~\cite{dasigi-etal-2021-dataset}.
    \item \textbf{Multiple sources:} the model needs to consider a collection of documents to determine which one is the best candidate to contain the correct answer (if any). Many open-domain QA resources fall in this category: e.g. MS MARCO~\cite{BajajCamposEtAl_2016_MS_MARCO_Human_Generated_MAchine_Reading_COmprehension_Dataset}, \anna{SearchQA~\cite{DunnSagunEtAl_2017_SearchQA_New_QA_Dataset_Augmented_with_Context_from_Search_Engine}} and TriviaQA~\cite{JoshiChoiEtAl_2017_TriviaQA_Large_Scale_Distantly_Supervised_Challenge_Dataset_for_Reading_Comprehension} come with retrieved Web snippets as the ``texts''. Similarly, some VQA datasets have multiple images as contexts~\cite{suhr-etal-2019-nlvr2}. 
    \item \textbf{Partial source:} The dataset provides documents that are necessary, but not sufficient to produce the correct answer. This may happen when the evidence snippets may be collected independently and not guaranteed to contain the answer, as in ARC~\cite{ClarkCowheyEtAl_2018_Think_you_have_Solved_Question_Answering_Try_ARC_AI2_Reasoning_Challenge}. Another frequent case is commonsense reasoning datasets such as RocStories~\cite{MostafazadehRothEtAl_2017_LSDSem_2017_Shared_Task_The_Story_Cloze_Test} or CosmosQA~\cite{HuangLeBrasEtAl_2019_Cosmos_QA_Machine_Reading_Comprehension_with_Contextual_Commonsense_Reasoning}: there is a text, and the correct answer depends on both the information in this text and implicit world knowledge. E.g for SemEval2018 Task 11~\cite{OstermannRothEtAl_2018_SemEval-2018_Task_11_Machine_Comprehension_Using_Commonsense_Knowledge} the organizers provided a commonsense reasoning dataset, and participants were free to use any external world knowledge resource. 
    \item \textbf{No sources.} The model needs to rely only on some external source of knowledge, such as the knowledge stored in the weights of a pre-trained language model, a knowledge base, or an information retrieval component. A notable example is commonsense reasoning datasets, such as Winograd Schema Challenge~\cite{LevesqueDavisEtAl_2012_Winograd_Schema_Challenge} or COPA~\cite{RoemmeleBejanEtAl_2011_Choice_of_Plausible_Alternatives_Evaluation_of_Commonsense_Causal_Reasoning,GordonKozarevaEtAl_2012_SemEval-2012_Task_7_Choice_of_Plausible_Alternatives_Evaluation_of_Commonsense_Causal_Reasoning}). 
\end{itemize*}

As shown in \autoref{fig:knowledge-sources}, this is also more of continuum than a strict taxonomy, \final{and some resources come with different versions of leaderboards based on how much evidence the model can access  \cite[e.g.][]{CiosiciCecilEtAl_2021_Perhaps_PTLMs_Should_Go_to_School_-_Task_to_Assess_Open_Book_and_Closed_Book_QA,TapaswiZhuEtAl_2016_MovieQA_Understanding_Stories_in_Movies_through_Question-Answering}}. As we go from a single well-matched source of knowledge to a large heterogeneous collection, the QA problem increasingly incorporates an element of information retrieval. The same could be said for long single sources, such as long documents or videos, if answers can be found in a single relevant excerpt and do not require a high-level summary of the whole context. So far, our QA/RC resources tend to target more complex reasoning for shorter texts, as it is more difficult to create difficult questions over larger contexts.

Arguably, an intermediate case between single-source and multiple-source cases are datasets that collect multiple sources per question, but provide them already coupled with questions, which turns each example into a single-source problem. For example, TriviaQA~\cite{JoshiChoiEtAl_2017_TriviaQA_Large_Scale_Distantly_Supervised_Challenge_Dataset_for_Reading_Comprehension} contains 95K questions, but 650K question-answer-evidence triplets.

\section{Conversational features}
\label{sec:discourse}

The vast majority of questions in datasets discussed so far were collected or created as standalone questions, targeting a \textit{static} source of evidence (text, knowledge base and/or any multimedia). The pragmatic context modeled in this setting is simply a set of standalone questions that could be asked in any order. But due to the active development of digital assistants, there is also active research on QA in conversational contexts: in addition to any sources of knowledge being discussed or used by the interlocutors, there is conversation history, which may be required to even interpret the question. For example, the question ``Where did Einstein die''? may turn into ``Where did he die?'' if it is a follow-up question; after that, the order of the questions can no longer be swapped. The key differences to the traditional RC setting is that (a) the conversation history grows dynamically as the conversation goes on, (b) it is \textit{not} the main source of information (that comes from some other context, a knowledge base, etc. \final{Most current conversational QA resources are text-based, but the multimodal setting is also possible \cite{LiLiEtAl_2022_MMCoQA_Conversational_Question_Answering_over_Text_Tables_and_Images}.}

While ``conversational QA'' may be intuitively associated with spoken (as opposed to written) language, the current resources for conversational QA do not necessarily originate in this way. For example, similarly to RC datasets like SQuAD, CoQA~\cite{ReddyChenEtAl_2019_CoQA_Conversational_Question_Answering_Challenge} was created in the written form, by crowd workers provided with prompts. It could be argued that the ``natural'' search engine queries have some spoken language features, but they also have their own peculiarities stemming from the fact that functionally, they are queries rather than questions (see \cref{sec:question-format}).

A big challenge in creating conversational datasets is making sure that the questions are really information-seeking rather than probing (\cref{sec:natural}), since humans would not normally use the latter with each other (except perhaps in language learning contexts or checking whether someone slept through a meeting). From the perspective of how much knowledge the questioner has, existing datasets can be grouped into three categories:

\begin{itemize*}

\item \textbf{Equal knowledge.} For example, CoQA~\cite{ReddyChenEtAl_2019_CoQA_Conversational_Question_Answering_Challenge} collected dialogues about the information in a passage (from seven domains) from two crowd workers, both of whom see the target passage. The interface discouraged the workers from using words occurring in the text.
\item \textbf{Unequal knowledge.} For example, QuAC~\cite{ChoiHeEtAl_2018_QuAC_Question_Answering_in_Context} is a collection of factual questions about a topic,\footnote{In most conversational QA datasets collected in the unequal-knowledge setup the target information is factual, and the simulated scenario is that only one of the participants has access to that information (but theoretically, anyone could have such access). An interesting alternative direction is questions where the other participant is the only possible source of information: personal questions. An example of that is CCPE-M~\cite{RadlinskiBalogEtAl_2019_Coached_conversational_preference_elicitation_case_study_in_understanding_movie_preferences}, a collection of dialogues where one party elicits the other party's movie preferences. \final{Another direction is investigating what knowledge the questioner already has: Curiosity \cite{RodriguezCrookEtAl_2020_Information_Seeking_in_Spirit_of_Learning_Dataset_for_Conversational_Curiosity} provides information-seeking dialogues annotated with the questioner's prior knowledge of entities mentioned (as well as dialogue act types and whether the questioners liked the answers they got).}}
asked by one crowdworker and answered by another (who has access to a Wikipedia article). A similar setup to QuAC was used for the Wizards of Wikipedia~\cite{DinanRollerEtAl_2018_Wizard_of_Wikipedia_Knowledge-Powered_Conversational_agents}, which, however, focuses on chitchat about Wikipedia topics rather than question answering, and could perhaps be seen as complementary to QuAC. ShARC~\cite{SaeidiBartoloEtAl_2018_Interpretation_of_Natural_Language_Rules_in_Conversational_Machine_Reading} uses more than two annotators for authoring the main and follow-up questions to simulate different stages of a dialogue. \anna{Arguably another case of unequal knowledge is when a human annotator writes questions in dialogue with a system generating the answers, as in ConvQuestions~\cite{ChristmannSahaRoyEtAl_2019_Look_before_you_Hop_Conversational_Question_Answering_over_Knowledge_Graphs_Using_Judicious_Context_Expansion}}.
\item \textbf{Repurposing ``natural'' dialogue-like data.} An example of this approach is Molweni~\cite{LiLiuEtAl_2020_Molweni_Challenge_Multiparty_Dialogues-based_Machine_Reading_Comprehension_Dataset_with_Discourse_Structure}, based on the Ubuntu Chat corpus. Its contribution is discourse level annotations in sixteen types of relations (comments, clarification questions, elaboration etc.) MANtIS~\cite{PenhaBalanEtAl_2019_Introducing_MANtIS_novel_Multi-Domain_Information_Seeking_Dialogues_Dataset} is similarly based on StackExchange dialogues, with a sample annotated for nine discourse categories. MSDialog~\cite{QuYangEtAl_2018_Analyzing_and_Characterizing_User_Intent_in_Information-seeking_Conversations} is based on Microsoft support forums, and the Ubuntu dialogue corpus~\cite{LowePowEtAl_2015_Ubuntu_Dialogue_Corpus_Large_Dataset_for_Research_in_Unstructured_MultiTurn_Dialogue_Systems} likewise contains many questions and answers from the Ubuntu ecosystem. 

\end{itemize*}

Again, these proposed distinctions are not clear-cut, and there are in-between cases. For instance, DoQA~\cite{CamposOtegiEtAl_2020_DoQA-Accessing_Domain-Specific_FAQs_via_Conversational_QA} is based on ``real information needs'' because the questions are based on StackExchange questions, but the actual questions were still generated by crowdworkers in the ``unequal knowledge'' scenario, with real queries serving as ``inspiration''. SHaRC~\cite{SaeidiBartoloEtAl_2018_Interpretation_of_Natural_Language_Rules_in_Conversational_Machine_Reading} has a separate annotation step in which crowd workers formulate a scenario in which the dialogue they see could take place, i.e. trying to reverse-engineer the information need.

An emerging area in conversational QA is question rewriting:\footnote{See also the task of ``decontextualization'' that could be used as``answer rewriting''~\cite{ChoiPalomakiEtAl_2021_Decontextualization_Making_Sentences_Stand-Alone}: in QA/RC, this means altering the sentences containing the answer so that they could be easily interpreted without reading the full text, e.g. by resolving coreference chains and replacing pronouns with nouns.} rephrasing questions in a way that would make them easier to be answered e.g. through Web search results. CANARD~\cite{ElgoharyPeskovEtAl_2019_Can_You_Unpack_That_Learning_to_Rewrite_Questions-in-Context} \anna{is a dataset of rewritten QuAC questions (50\% of the original QuAC)}, and SaAC~\cite{AnanthaVakulenkoEtAl_2020_Open-Domain_Question_Answering_Goes_Conversational_via_Question_Rewriting} is similarly based on a collection of TREC resources. \anna{QReCC~\cite{AnanthaVakulenkoEtAl_2020_Open-Domain_Question_Answering_Goes_Conversational_via_Question_Rewriting} is a dataset of dialogues with seed questions from QuAC \cite{ReddyChenEtAl_2019_CoQA_Conversational_Question_Answering_Challenge}, Natural Questions \cite{KwiatkowskiPalomakiEtAl_2019_Natural_Questions_Benchmark_for_Question_Answering_Research}, and TREC CAst \cite{JeffreyChenyanEtAl_2019_CAsT_2019_conversational_assistance_track_overview}}, with follow-up questions written by professional annotators. All questions come in two versions: the ``natural'' and search-engine-friendly version, e.g. by resolving pronouns to the nouns mentioned in the dialogue history. Disfl-QA~\cite{gupta2021disflqa} is a derivative of SQuAD with questions containing typical conversational ``disfluencies'' such as ``uh'' and self-corrections.

The above line of work is what one could call \textit{conversational QA}. In parallel with that, there are datasets for \textit{dialogue comprehension}, i.e. datasets for testing the ability to understand dialogues as opposed to static texts. They are ``probing'' in the same sense as e.g. RACE~\cite{LaiXieEtAl_2017_RACE_Large-scale_ReAding_Comprehension_Dataset_From_Examinations}: the only difference is that the text is a dialogue script. In this category, FriendsQA~\cite{YangChoi_2019_FriendsQA_Open-Domain_Question_Answering_on_TV_Show_Transcripts} is based on transcripts of the `Friends' TV show, with questions and extractive answers generated by crowd workers. There is also a Cloze-style dataset based on the same show~\cite{MaJurczykEtAl_2018_Challenging_Reading_Comprehension_on_Daily_Conversation_Passage_Completion_on_Multiparty_Dialog}, targeting named entities. \final{QAConv \cite{WuMadottoEtAl_2022_QAConv_Question_Answering_on_Informative_Conversations} targets informative dialogues, such as work channels.}  DREAM~\cite{SunYuEtAl_2019_DREAM_Challenge_Data_Set_and_Models_for_Dialogue-Based_Reading_Comprehension} is a multi-choice dataset based on English exam data, with texts being dialogues. 

Another related subfield is task-oriented (also known as goal-oriented) dialogue, which typically includes questions as well as transactional operations. The goal is for the user to collect the information they need and then perform a certain action (e.g. find out what flights are available, choose and book one). There is some data for conversations with travel agents~\cite{SRIInternational_2011_SRIs_Amex_Travel_Agent_Data,KimDHaroEtAl_2016_Dialog_State_Tracking_Challenge_5_Handbook_v31}, conducting meetings~\cite{AlexanderssonBuschbeck-WolfEtAl_1998_Dialogue_acts_in_Verbmobil_2}, navigation, scheduling and weather queries to an in-car personal assistant~\cite{EricManning_2017_Key-Value_Retrieval_Networks_for_Task-Oriented_Dialogue}, and other~\cite{SerbanLoweEtAl_2015_Survey_of_Available_Corpora_for_Building_DataDriven_Dialogue_Systems,AsriSchulzEtAl_2017_Frames_Corpus_for_Adding_Memory_to_Goal-Oriented_Dialogue_Systems}, as well as multi-domain resources~\cite{BudzianowskiWenEtAl_2018_MultiWOZ_-_Large-Scale_Multi-Domain_Wizard-of-Oz_Dataset_for_Task-Oriented_Dialogue_Modelling,PeskovClarkeEtAl_2019_Multi-Domain_Goal-Oriented_Dialogues_MultiDoGO_Strategies_toward_Curating_and_Annotating_Large_Scale_Dialogue_Data}.

Conversational QA is also actively studied in information retrieval, and that community has produced many insights about actual human behavior in information-seeking dialogues. For instance, outside of maybe conference poster sessions and police interrogations, dialogues do not usually consist only of questions and answers, which is e.g. the CoQA setting. Studies of human-system interaction \cite[e.g.][]{TrippasSpinaEtAl_2018_Informing_Design_of_Spoken_Conversational_Search_Perspective_Paper} elaborate on the types of conversational moves performed by the users (such as informing, rejecting, promising etc.) and how they could be modeled. In conversational QA there are also potentially many more signals useful in evaluation than simply correctness: e.g. MISC~\cite{ThomasMcDuffEtAl_2017_MISC_data_set_of_information-seeking_conversations} is a small-scale resource produced by in-house MS staff that includes not only transcripts, but also audio, video, affectual and physiological signals, as well as recordings of search and other computer use and post-task surveys on emotion, success, and effort.

\section{Domains}
\label{sec:domains}

One major source of confusion in the domain adaptation literature is the very notion of ``domain'', which is often used to mean the source of data rather than any coherent criterion such as topic, style, genre, or linguistic register~\cite{RamponiPlank_2020_Neural_Unsupervised_Domain_Adaptation_in_NLP-A_Survey}. In the current QA/RC literature it seems to be predominantly used in the senses of ``topic'' and ``genre'' (a type of text, with a certain structure, stylistic conventions, and area of use). For instance, one could talk about the domains of programming or health, but either of them could be the subject of forums, encyclopedia articles, etc. which are ``genres'' in the linguistic sense. The below classification is primarily based on the understanding of ``domain'' as ``genre''. %

\textbf{Encyclopedia.} %
Wikipedia is probably the most widely used source of knowledge for constructing QA/RC datasets \cite[e.g.][]{HewlettLacosteEtAl_2016_WikiReading_Novel_Large-scale_Language_Understanding_Task_over_Wikipedia,YangYihEtAl_2015_WikiQA_A_Challenge_Dataset_for_Open-Domain_Question_Answering,RajpurkarZhangEtAl_2016_SQuAD_100000+_Questions_for_Machine_Comprehension_of_Text,DuaWangEtAl_2019_DROP_Reading_Comprehension_Benchmark_Requiring_Discrete_Reasoning_Over_Paragraphs}. The QA resources of this type, together with those based on knowledge bases and Web snippets, constitute what is in some communities referred to as ``open-domain'' QA\footnote{In other communities, ``open-domain'' somewhat confusingly implies not something about a ``domain'' per se, but a format: that no evidence is given for a question, and that information must be retrieved from some corpus, which is often Wikipedia.}. Note that here the term ``domain'' is used in the ``topic'' sense: Wikipedia, as well as Web and knowledge bases, contain much specialist knowledge, and the difference from the resources described below as ``expert materials'' is only that it is not restricted to particular topics.

\textbf{Fiction.} 
While fiction is one of the areas where large amounts of public-domain data is available, surprisingly few attempts were made to use them as reading comprehension resources, perhaps due to the incentive for more ``useful'' information-seeking QA work. CBT~\cite{HillBordesEtAl_2015_Goldilocks_Principle_Reading_Childrens_Books_with_Explicit_Memory_Representations} is an early and influential Cloze dataset based on children's stories. BookTest~\cite{BajgarKadlecEtAl_2017_Embracing_data_abundance_BookTest_Dataset_for_Reading_Comprehension} expands the same methodology to a larger number of project Gutenberg books. Being Cloze datasets, they inherit the limitations of the format discussed in \cref{sec:format-cloze}. \final{FairyTaleQA \cite{XuWangEtAl_2022_Fantastic_Questions_and_Where_to_Find_Them_FairytaleQA_-_Authentic_Dataset_for_Narrative_Comprehension} is a recent multi-choice dataset based on academic tests.} A key challenge of fiction is understanding a long text; this challenge is addressed by NarrativeQA~\cite{KociskySchwarzEtAl_2018_NarrativeQA_Reading_Comprehension_Challenge} and QuALITY~\cite{PangParrishEtAl_2022_QuALITY_Question_Answering_with_Long_Input_Texts_Yes}. %

The above resources target literary or genre fiction: long, complex narratives created for human entertainment or instruction. NLP papers also often rely on fictional mini-narratives written by crowdworkers for the purpose of RC tests. Examples of this genre include MCTest~\cite{RichardsonBurgesEtAl_2013_MCTest_A_Challenge_Dataset_for_the_Open-Domain_Machine_Comprehension_of_Text}, MCScript~\cite{OstermannRothEtAl_2018_SemEval-2018_Task_11_Machine_Comprehension_Using_Commonsense_Knowledge,ostermann-etal-2018-mcscript,modi-etal-2016-inscript}, and RocStories~\cite{MostafazadehRothEtAl_2017_LSDSem_2017_Shared_Task_The_Story_Cloze_Test}.

\textbf{Academic tests.} 
This is one of the few ``genres'' where experts devise high-quality discriminative probing %
questions. Most of the current datasets were sourced from materials written by expert teachers to test students, which in addition to different subjects yields the ``natural'' division by student level (different school grades, college etc.). Arguably, it corresponds to level of difficulty of target concepts (if not necessarily language). Among the college exam resources, CLEF competitions~\cite{PenasUngerEtAl_2014_Overview_of_CLEF_Question_Answering_Track_2014,PenasUngerEtAl_2015_Overview_of_the_CLEF_Question_Answering_Track_2015} and NTCIR QA Lab~\cite{ShibukiSakamotoEtAl_2014_Overview_of_NTCIR-11_QA-Lab_Task} were based on small-scale data from Japanese university entrance exams. RACE-C~\cite{LiangLiEtAl_2019_new_multi-choice_reading_comprehension_dataset_for_curriculum_learning} draws on similar data developed for Chinese university admissions. 
ReClor~\cite{YuJiangEtAl_2019_ReClor_Reading_Comprehension_Dataset_Requiring_Logical_Reasoning} is a collection of reading comprehension questions from standartized admission tests like GMAT and LSAT, selected specifically to target logical reasoning. 

In the school-level tests, the most widely-used datasets are RACE~\cite{LaiXieEtAl_2017_RACE_Large-scale_ReAding_Comprehension_Dataset_From_Examinations} and DREAM~\cite{SunYuEtAl_2019_DREAM_Challenge_Data_Set_and_Models_for_Dialogue-Based_Reading_Comprehension}, both comprised of tests created by teachers for testing the reading comprehension of English by Chinese students (on narratives and multi-party dialogue transcripts, respectively). ARC~\cite{ClarkCowheyEtAl_2018_Think_you_have_Solved_Question_Answering_Try_ARC_AI2_Reasoning_Challenge} targets science questions authored for US school tests. OpenBookQA~\cite{MihaylovClarkEtAl_2018_Can_Suit_of_Armor_Conduct_Electricity_New_Dataset_for_Open_Book_Question_Answering} also targets elementary science knowledge, but the questions were written by crowdworkers. ProcessBank~\cite{BerantSrikumarEtAl_2014_Modeling_Biological_Processes_for_Reading_Comprehension} is a small-scale multi-choice dataset based on biology textbooks.

\final{\textbf{Trivia.} Resources based on human knowledge competitions overlap with encyclopedia in subject matter, but this is a separate genre: the questions are authored by domain experts specifically to be discriminative tests of human knowledge, and, unlike in academic tests, the participants engage in the QA activity for fun. Examples include TriviaQA~\cite{JoshiChoiEtAl_2017_TriviaQA_Large_Scale_Distantly_Supervised_Challenge_Dataset_for_Reading_Comprehension}, Quizbowl \cite{Boyd-GraberFengEtAl_2018_Human-Computer_Question_Answering_Case_for_Quizbowl,RodriguezFengEtAl_2021_Quizbowl_Case_for_Incremental_Question_Answering} and Jeopardy\footnote{Many versions of question sets from Jeopardy show are currently circulated, e.g. \url{https://huggingface.co/datasets/jeopardy}}. See \cite{Boyd-Graber_2019_What_Question_Answering_can_Learn_from_Trivia_Nerds} for the discussion of what makes questions discriminative, and~\cite{RodriguezBoyd-Graber_2021_Evaluation_Paradigms_in_Question_Answering} for how that should be taken into account in NLP leaderboards.}

\textbf{News.} 
Given the increasing problem of online misinformation (see \cref{sec:missing-data}), QA for news is a highly societally important area of research, but it is hampered by the lack of public-domain data%
. The best-known reading comprehension dataset based on news is undoubtedly the CNN/Daily Mail Cloze dataset~\cite{HermannKociskyEtAl_2015_Teaching_Machines_to_Read_and_Comprehend}, focusing on the understanding of named entities and coreference relations within a text. Subsequently NewsQA~\cite{trischler-etal-2017-newsqa} also relied on CNN data; it is an extractive dataset with questions written by crowd workers. Most recently, NLQuAD~\cite{soleimani-etal-2021-nlquad} is an extractive benchmark with ``non-factoid'' questions (originally BBC news article subheadings) that need to be matched with longer spans within the articles. In multi-choice format, a section of QuAIL~\cite{RogersKovalevaEtAl_2020_Getting_Closer_to_AI_Complete_Question_Answering_Set_of_Prerequisite_Real_Tasks} is based on CC-licensed news. There is also a small test dataset of temporal questions for news events over a New York Times archive~\cite{WangJatowtEtAl_2021_Improving_question_answering_for_event-focused_questions_in_temporal_collections_of_news_articles}, and a larger resource for QA over historical news collections \cite{WangJatowtEtAl_2022_ArchivalQA_Large-scale_Benchmark_Dataset_for_Open_Domain_Question_Answering_over_Historical_News_Collections}.

\textbf{E-commerce.} This category focuses on the genre of product reviews. Two such resources are based on Amazon review data: one was sourced from a Web crawl of questions and answers about products posed by users~\cite{McAuleyYang_2016_Addressing_Complex_and_Subjective_Product-Related_Queries_with_Customer_Reviews}, and the more recent one (AmazonQA~\cite{GuptaKulkarniEtAl_2019_AmazonQA_Review-Based_Question_Answering_Task}) built upon it by cleaning up the data, and providing review snippets and (automatic) answerability annotation. SubjQA~\cite{BjervaBhutaniEtAl_2020_SubjQA_Dataset_for_Subjectivity_and_Review_Comprehension} is based on reviews from more sources than just Amazon, has manual answerability annotation and, importantly, is the first QA dataset to also include labels for subjectivity of answers.

\textbf{Expert materials.} This is a loose group that could be further subdivided by the genre of its sources (manuals, reports, scientific papers etc.). What they have in common is that the topic is narrow and only known to experts. %
Most existing resources are based on answers provided by volunteer experts: e.g. TechQA~\cite{CastelliChakravartiEtAl_2020_TechQA_Dataset} is based on naturally-occurring questions from tech forums. A less common option is to hire experts, as done for Qasper~\cite{dasigi-etal-2021-dataset}: a dataset of expert-written questions over NLP papers. The ``volunteer expert'' setting is the focus of the subfield of \textit{community QA}. It deserves a separate survey, but the key difference to the ``professional'' support resources is that the answers are provided by volunteers with varying levels of expertise, on platforms such as WikiAnswers~\cite{AbujabalRoyEtAl_2019_ComQA_Community-sourced_Dataset_for_Complex_Factoid_Question_Answering_with_Paraphrase_Clusters}, Reddit~\cite{FanJerniteEtAl_2019_ELI5_Long_Form_Question_Answering}, or AskUbuntu~\cite{dos-santos-etal-2015-learning}. Since the quality and amount of both questions and answers vary a lot, new QA subtasks emerged, including duplicate question detection and ranking multiple answers for the same question~\cite{NakovMarquezEtAl_2015_SemEval-2015_Task_3_Answer_Selection_in_Community_Question_Answering,NakovMarquezEtAl_2016_SemEval-2016_Task_3_Community_Question_Answeringb,NakovHoogeveenEtAl_2017_SemEval-2017_Task_3_Community_Question_Answering}.

One of the subject areas for which there are many expert-curated QA/RC resources is \textit{biomedical QA.}  BioASQ is a small-scale biomedical corpus targeting different NLP system capabilities (boolean questions, concept retrieval, text retrieval), that were initially formulated by experts as a part of CLEF competitions~\cite{PenasUngerEtAl_2014_Overview_of_CLEF_Question_Answering_Track_2014,PenasUngerEtAl_2015_Overview_of_the_CLEF_Question_Answering_Track_2015,TsatsaronisBalikasEtAl_2015_overview_of_BIOASQ_large-scale_biomedical_semantic_indexing_and_question_answering_competition}. \final{Mash-QA \cite{ZhuAhujaEtAl_2020_Question_Answering_with_Long_Multiple-Span_Answers} poses the challenge of answering healthcare questions by extracting multiple relevant spans. MIMICSQL \cite{WangShiEtAl_2020_Text-to-SQL_Generation_for_Question_Answering_on_Electronic_Medical_Records} is a dataset of SQL queries for a medical records database, paired with equivalent natural language questions (machine-generated and human-paraphrased).}  PubMedQA~\cite{JinDhingraEtAl_2019_PubMedQA_Dataset_for_Biomedical_Research_Question_Answering} is a corpus of biomedical literature abstracts that treats titles of articles as pseudo-questions, most of the abstract as context, and the final sentence of the abstract as the answer (with a small manually labeled section and larger unlabeled/artificially labeled section). In the healthcare area, CliCR~\cite{SusterDaelemans_2018_CliCR_Dataset_of_Clinical_Case_Reports_for_Machine_Reading_Comprehension} is a Cloze-style dataset of clinical records, and Head-QA~\cite{VilaresGomez-Rodriguez_2019_HEAD-QA_Healthcare_Dataset_for_Complex_Reasoning} is a multimodal multi-choice dataset written to test human experts in medicine, chemistry, pharmacology, psychology, biology, and nursing. emrQA~\cite{PampariRaghavanEtAl_2018_emrQA_Large_Corpus_for_Question_Answering_on_Electronic_Medical_Records} is an extractive dataset of clinical records with questions generated from templates, repurposing annotations from other NLP tasks such as NER. There is also data specifically on the COVID pandemic~\cite{MollerReinaEtAl_2020_COVID-QA_Question_Answering_Dataset_for_COVID-19}.

\textbf{Social media.} 
Social media data present a unique set of challenges: the user speech is less formal, more likely to contain typos and misspellings, and more likely to contain platform-specific phenomena such as hashtags and usernames. So far there are not so many such resources, but one example is  TweetQA~\cite{XiongWuEtAl_2019_TWEETQA_Social_Media_Focused_Question_Answering_Dataset}, which crowdsourced questions and answers for (news-worthy) tweet texts. 

\textbf{Multi-domain.} 
\final{It is a well established result in NLP research that success on one dataset does not necessarily transfer to another, and this concerns the field of QA/RC as well \cite{Yatskar_2019_Qualitative_Comparison_of_CoQA_SQuAD_20_and_QuAC,AkulaChangpinyoEtAl_2021_CrossVQA_Scalably_Generating_Benchmarks_for_Systematically_Testing_VQA_Generalization}. %
The issue is even more pronounced in cross-domain setting.} However, so far there are very few attempts to create multi-domain datasets that could encourage generalization by design, and, as discussed above, they are not necessarily based on the same notion of ``domain''. In the sense of ``genre'', the first one was CoQA~\cite{ReddyChenEtAl_2019_CoQA_Conversational_Question_Answering_Challenge}, combining prompts from children’s stories, fiction, high school English exams, news articles, Wikipedia, science and Reddit articles. It was followed by QuAIL~\cite{RogersKovalevaEtAl_2020_Getting_Closer_to_AI_Complete_Question_Answering_Set_of_Prerequisite_Real_Tasks}, a multi-choice dataset balanced across news, fiction, user stories and blogs. 

In the sense of ``topic'', two more datasets are presented as ``multi-domain'': MMQA~\cite{GuptaKumariEtAl_2018_MMQA_Multi-domain_Multi-lingual_Question-Answering_Framework_for_English_and_Hindi} is an English-Hindi dataset of Web articles that is presented as a multi-domain dataset, but is based on Web articles on the topics of tourism, history, diseases, geography, economics, and environment. In the same vein, MANtIS~\cite{PenhaBalanEtAl_2019_Introducing_MANtIS_novel_Multi-Domain_Information_Seeking_Dialogues_Dataset} is a collection of information-seeking dialogues from StackExchange fora across 14 topics (Apple, AskUbuntu, DBA, DIY, ELectronics, English, Gaming, GIS, Physics, Scifi, Security, Stats, Travel, World-building). %

There are also ``collective'' datasets, formed as a collection of existing datasets, which may count as ``multi-domain'' by different criteria. In the sense of ``genre'', ORB~\cite{DuaGottumukkalaEtAl_2019_ORB_Open_Reading_Benchmark_for_Comprehensive_Evaluation_of_Machine_Reading_Comprehension} includes data based on news, Wikipedia, fiction. MultiReQA~\cite{GuoYangEtAl_2020_MultiReQA_Cross-Domain_Evaluation_for_Retrieval_Question_Answering_Models} comprises 8 datasets, targeting textbooks, Web snippets, Wikipedia, scientific articles.

\section{Languages}
\label{sec:languages}
\subsection{Monolingual resources}
\label{sec:languages-mono}

As in other areas of NLP, the ``default'' language of QA and RC is \textbf{English}~\cite{Bender_2019_BenderRule_On_Naming_Languages_We_Study_and_Why_It_Matters}, and most of this survey discusses English resources. The second best-resourced language in terms or QA/RC data is \textbf{Chinese}, which has the counterparts of many popular English resources. Besides SQuAD-like resources~\cite{CuiLiuEtAl_2019_Span-Extraction_Dataset_for_Chinese_Machine_Reading_Comprehension,ShaoLiuEtAl_2019_DRCD_Chinese_Machine_Reading_Comprehension_Dataset}, there is shared task data for open-domain QA based on structured and text data~\cite{DuanTang_2018_Overview_of_NLPCC_2017_Shared_Task_Open_Domain_Chinese_Question_Answering}, as well as a resource for table QA \cite{SunYangEtAl_2020_TableQA_Large-Scale_Chinese_Text-to-SQL_Dataset_for_Table-Aware_SQL_Generation}. 
WebQA is an open-domain dataset of community questions with entities as answers, and web snippets annotated for whether they provide the correct answer~\cite{LiLiEtAl_2016_Dataset_and_Neural_Recurrent_Sequence_Labeling_Model_for_Open-Domain_Factoid_Question_Answering}. ReCO~\cite{WangYaoEtAl_2020_ReCO_Large_Scale_Chinese_Reading_Comprehension_Dataset_on_Opinion} targets boolean questions from user search engine queries. There are also cloze-style datasets based on news, fairy tales, and children's reading material, mirroring CNN/Daily Mail and CBT~\cite{CuiLiuEtAl_2016_Consensus_Attention-based_Neural_Networks_for_Chinese_Reading_Comprehension,CuiLiuEtAl_2018_Dataset_for_First_Evaluation_on_Chinese_Machine_Reading_Comprehension}, as well as a recent sentence-level cloze resource~\cite{CuiLiuEtAl_2020_Sentence_Cloze_Dataset_for_Chinese_Machine_Reading_Comprehension}.  
DuReader~\cite{HeLiuEtAl_2018_DuReader_Chinese_Machine_Reading_Comprehension_Dataset_from_Real-world_Applications} is a freeform QA resource based on search engine queries and community QA. In terms of niche topics, there are Chinese datasets focusing on history textbooks~\cite{ZhangZhao_2018_One-shot_Learning_for_Question-Answering_in_Gaokao_History_Challenge}, biomedical exams \cite{LiZhongEtAl_2021_MLEC-QA_Chinese_Multi-Choice_Biomedical_Question_Answering_Dataset}, and maternity forums~\cite{XuPeiEtAl_2020_MATINF_Jointly_Labeled_Large-Scale_Dataset_for_Classification_Question_Answering_and_Summarization}.

In the third place we have \textbf{Russian}, which a version of SQuAD~\cite{EfimovChertokEtAl_2020_SberQuAD_-_Russian_Reading_Comprehension_Dataset_Description_and_Analysis}, a dataset for open-domain QA over Wikidata~\cite{KorablinovBraslavski_2020_RuBQ_Russian_Dataset_for_Question_Answering_over_Wikidata}, a boolean QA dataset~\cite{GlushkovaMachnevEtAl_2020_DaNetQA_yesno_Question_Answering_Dataset_for_Russian_Language}, and datasets for cloze-style commonsense reasoning and multi-choice, multi-hop RC ~\cite{FenogenovaMikhailovEtAl_2020_Read_and_Reason_with_MuSeRC_and_RuCoS_Datasets_for_Machine_Reading_Comprehension_for_Russian}. 

The fourth best resourced language is \textbf{Japanese}, with a Cloze RC dataset \cite{WataraiTsuchiya_2020_Developing_Dataset_of_Japanese_Slot_Filling_Quizzes_Designed_for_Evaluation_of_Machine_Reading_Comprehension}, a manual translation of a part of SQuAD \cite{AsaiEriguchiEtAl_2018_Multilingual_Extractive_Reading_Comprehension_by_Runtime_Machine_Translation}, and a commonsense reasoning resource~\cite{OmuraKawaharaEtAl_2020_Method_for_Building_Commonsense_Inference_Dataset_based_on_Basic_Events}.

Three more languages have their versions of SQuAD~\cite{RajpurkarZhangEtAl_2016_SQuAD_100000+_Questions_for_Machine_Comprehension_of_Text}: \textbf{French}~\cite{KeraronLancrenonEtAl_2020_Project_PIAF_Building_Native_French_Question-Answering_Dataset,dHoffschmidtVidalEtAl_2020_FQuAD_French_Question_Answering_Dataset}, \textbf{Vietnamese}~\cite{NguyenNguyenEtAl_2020_Vietnamese_Dataset_for_Evaluating_Machine_Reading_Comprehension}, and \textbf{Korean}~\cite{LimKimEtAl_2019_KorQuAD10_Korean_QA_Dataset_for_Machine_Reading_Comprehension}, and there are three more small-scale evaluation sets (independently collected for \textbf{Arabic}~\cite{MozannarMaamaryEtAl_2019_Neural_Arabic_Question_Answering}), human-translated to \textbf{French}~\cite{AsaiEriguchiEtAl_2018_Multilingual_Extractive_Reading_Comprehension_by_Runtime_Machine_Translation}). \textbf{Polish} has a small dataset of open-domain questions based on Wikipedia ``Did you know...?'' data~\cite{MarcinczukPtakEtAl_2013_Open_dataset_for_development_of_Polish_question_answering_systems}. And, to the best of our knowledge, this is it: not even the relatively well-resourced languages like German necessarily have any monolingual QA/RC data. There is more data for individual languages that is part of multilingual benchmarks, but that comes with a different set of issues (\cref{sec:languages-multi}).

In the absence of data, the researchers resort to machine translation of English resources. For instance, there is such SQuAD data for \textbf{Spanish}~\cite{CarrinoCosta-jussaEtAl_2019_Automatic_Spanish_Translation_of_SQuAD_Dataset_for_Multilingual_Question_Answering}, \textbf{Arabic}~\cite{MozannarMaamaryEtAl_2019_Neural_Arabic_Question_Answering}, \textbf{Italian}~\cite{CroceZelenanskaEtAl_2019_Enabling_deep_learning_for_large_scale_question_answering_in_Italian}, \textbf{Korean}~\cite{LeeYoonEtAl_2018_Semi-supervised_Training_Data_Generation_for_Multilingual_Question_Answering}. However, this has clear limitations: machine translation comes with its own problems and artifacts, and in terms of content even the best translations could differ from the questions that would be ``naturally'' asked by the speakers of different languages.

The fact that so few languages have many high-quality QA/RC resources reflecting the idiosyncrasies and information needs of the speakers of their languages says a lot about the current distribution of funding for data development, and the NLP community appetite for publishing non-English data at top NLP conferences. There are reports of reviewer bias~\cite{RogersAugenstein_2020_What_Can_We_Do_to_Improve_Peer_Review_in_NLP}: such work may be perceived as ``niche'' and low-impact, which makes it look like a natural candidate for second-tier venues\footnote{e.g. \textit{Findings of EMNLP} was specifically created as a venue for which ``there is no requirement for high perceived impact, and accordingly solid work in untrendy areas and other more niche works will be eligible'' (\url{}https://2020.emnlp.org/blog/2020-04-19-findings-of-emnlp)}, which makes such work hard to pursue for early career researchers. 

This situation is not only problematic in terms of inclusivity and diversity (where it contributes to unequal access to the latest technologies around the globe). The focus on English is also counter-productive because it creates the wrong impression of progress on QA/RC vs the \textit{subset} of QA/RC that is easy in English. For instance, as pointed out by the authors of TydiQA~\cite{ClarkChoiEtAl_2020_TyDi_QA_Benchmark_for_Information-Seeking_Question_Answering_in_Typologically_Diverse_Languages}, questions that can be solved by string matching are easy in English (a morphologically poor language), but can be very difficult in languages with many morphophonological alternations and compounding. 

Another factor contributing to the perception of non-English work as ``niche'' and low-impact is that many such resources are ``replications'' of successful English resources, which makes them look derivative (see e.g. the above-mentioned versions of SQuAD). However, conceptually the contribution of such work is arguably comparable to incremental modifications of popular NLP architectures (a genre that does not seem to raise objections of low novelty), while having potentially much larger real-world impact. Furthermore, such work may also require non-trivial adaptations to transfer an existing methodology to a different language, and/or propose first-time innovations. For instance, MATINF~\cite{XuPeiEtAl_2020_MATINF_Jointly_Labeled_Large-Scale_Dataset_for_Classification_Question_Answering_and_Summarization} is a Chinese dataset jointly labeled for classification, QA and summarization, so that the same data could be used to train for all three tasks. The contribution of ~\citet{WataraiTsuchiya_2020_Developing_Dataset_of_Japanese_Slot_Filling_Quizzes_Designed_for_Evaluation_of_Machine_Reading_Comprehension} is not merely a Japanese version of CBT, but also a methodology to overcome some of its limitations.

\subsection{Multilingual resources}
\label{sec:languages-multi}

One way in which non-English work seems to be easier to publish is multilingual resources. Some of them are data from cross-lingual shared tasks\footnote{See e.g. QALD-4.1~\cite{PenasUngerEtAl_2014_Overview_of_CLEF_Question_Answering_Track_2014}, IJCNLP-2017 Task 5~\cite{GuoLiuEtAl_2017_IJCNLP-2017_Task_5_Multi-choice_Question_Answering_in_Examinations}.
}, and also independent academic resources (such as English-Chinese cloze-style XCMRC~\cite{LiuDengEtAl_2019_XCMRC_Evaluating_Cross-Lingual_Machine_Reading_Comprehension}). But in terms of number of languages, the spotlight is currently on the following larger-scale resources:

\begin{itemize*}
    \item MLQA~\cite{LewisOguzEtAl_2020_MLQA_Evaluating_Cross-lingual_Extractive_Question_Answering} targets extractive QA over Wikipedia with partially parallel texts in seven languages: English, Arabic, German, Spanish, Hindi, Vietnamese and Simplified Chinese. The questions are crowdsourced and translated. 
    \item XQuAD~\cite{ArtetxeRuderEtAl_2020_On_Cross-lingual_Transferability_of_Monolingual_Representations} is a subset of SQuAD professionally translated into 10 languages: Spanish, German, Greek, Russian, Turkish, Arabic, Vietnamese, Thai, Chinese, and Hindi.
    \item XQA~\cite{LiuLinEtAl_2019_XQA_Cross-lingual_Open-domain_Question_Answering_Dataset} is an open-domain QA dataset targeting entities; it provides training data for English, and test and development data for English and eight other languages: French, German, Portuguese, Polish, Chinese, Russian, Ukrainian, and Tamil.
    \item TydiQA~\cite{ClarkChoiEtAl_2020_TyDi_QA_Benchmark_for_Information-Seeking_Question_Answering_in_Typologically_Diverse_Languages} is the first resource of ``natural'' factoid questions in ten typologically diverse languages in addition to English: Arabic, Bengali, Finnish, Japanese, Indonesian, Kiswahili, Korean, Russian, Telugu, and Thai. 
    \item XOR QA~\cite{AsaiKasaiEtAl_2020_XOR_QA_Cross-lingual_Open-Retrieval_Question_Answering} builds on Tidy QA data to pose the task of cross-lingual QA: answering questions, where the answer data is unavailable in the same language as the question. It is a subset of TidyQA with data in seven languages: Arabic, Bengali, Finnish, Japanese, Korean, Russian and Telugu, with English as the ``pivot'' language (professionally translated). 
    \item XQuAD-R and MLQA-R~\cite{RoyConstantEtAl_2020_LAReQA_Language-agnostic_answer_retrieval_from_multilingual_pool} are based on the above-mentioned XQuAD and MLQA extractive QA resources, recast as multilingual information retrieval tasks.
    \item MKQA~\cite{LongpreLuEtAl_2020_MKQA_Linguistically_Diverse_Benchmark_for_Multilingual_Open_Domain_Question_Answering} is based on professional translations of a subset of Natural Questions~\cite{KwiatkowskiPalomakiEtAl_2019_Natural_Questions_Benchmark_for_Question_Answering_Research}, professionally translated into 26 languages, focusing on ``translation invariant'' questions.
\end{itemize*}

While these resources are a very valuable contribution, in multilingual NLP they seem to be playing the role similar to the role that the large-scale language models play in development of NLP models: the small labs are effectively out of the competition~\cite{Rogers_2019_How_Transformers_broke_NLP_leaderboards}. In comparison with large multilingual leaderboards, monolingual resources are perceived as ``niche'', less of a valuable contribution, less deserving of the main track publications on which careers of early-stage researchers depend. But such scale is only feasible for industry-funded research: of all the above multilingual datasets, only the smallest one (XQA) was not produced in affiliation with either Google, Apple, or Facebook. A recent academic effort is xGQA, posing the task of multilingual \textit{and} multimodal QA \cite{PfeifferGeigleEtAl_2022_xGQA_Cross-Lingual_Visual_Question_Answering}; it provides human-authored translations only for the test set of GQA \cite{HudsonManning_2019_GQA_New_Dataset_for_Real-World_Visual_Reasoning_and_Compositional_Question_Answering}. 

Furthermore, scale is not necessarily the best answer: focus on multilinguality necessarily requires missing a lot of nuance that is only possible for in-depth work on individual languages performed by experts in those languages. A key issue in multilingual resources is collecting data that is homogeneous enough across languages to be considered a fair and representative cross-lingual benchmark. That objective is necessarily competing with the objective of getting a natural and representative sample of questions in each individual language. To prioritize the latter objective, we would need comparable corpora of naturally occurring multilingual data. This is what happened in XQA~\cite{LiuLinEtAl_2019_XQA_Cross-lingual_Open-domain_Question_Answering_Dataset} (based on the ``Did you know... ?'' Wikipedia question data), but there is not much such data that is in public domain. Tidy QA~\cite{ClarkChoiEtAl_2020_TyDi_QA_Benchmark_for_Information-Seeking_Question_Answering_in_Typologically_Diverse_Languages} attempts to approximate ``natural'' questions by prompting speakers to formulate questions for the topics, on which they are shown the header excerpts of Wikipedia articles, but it is hard to tell to what degree this matches real information needs, or samples all the linguistic phenomena that are generally prominent in questions for this language and should be represented.%

A popular solution that sacrifices representativeness of individual languages for cross-lingual homogeneity is using translation, as it was done in MLQA~\cite{LewisOguzEtAl_2020_MLQA_Evaluating_Cross-lingual_Extractive_Question_Answering}, xQuaD~\cite{ArtetxeRuderEtAl_2020_On_Cross-lingual_Transferability_of_Monolingual_Representations}, and MKQA~\cite{LongpreLuEtAl_2020_MKQA_Linguistically_Diverse_Benchmark_for_Multilingual_Open_Domain_Question_Answering}. However, translationese has many issues. In addition to the high cost, even the best human translation is not necessarily similar to naturally occurring question data, since languages differ in what information is made explicit or implicit~\cite{ClarkChoiEtAl_2020_TyDi_QA_Benchmark_for_Information-Seeking_Question_Answering_in_Typologically_Diverse_Languages}, and cultures also differ in what kinds of questions typically get asked. 

A separate (but related) problem is that it is also not guaranteed that translated questions will have answers in the target language data. This issue lead XQuAD to translating both questions and texts, MLQA -- to partial cross-lingual coverage, MKQA -- to providing only questions and answers, without the evidence texts, and XOR QA ~\cite{AsaiKasaiEtAl_2020_XOR_QA_Cross-lingual_Open-Retrieval_Question_Answering} -- to positing the task of cross-lingual QA.

One more issue in multilingual NLP that does not seem to have received much attention in QA/RC research is code-switching~\cite{SitaramChanduEtAl_2020_Survey_of_Code-switched_Speech_and_Language_Processing}, even though it clearly has a high humanitarian value. For instance, in the US context better question answering with code-switched English/Spanish data could be highly useful in the civil service and education, supporting the learning of immigrant children and social integration of their parents. So far there are only a few small-scale resources for Hindi~\cite{RaghaviChinnakotlaEtAl_2015_Answer_ka_type_kya_he_Learning_to_Classify_Questions_in_Code-Mixed_Language,GuptaChinnakotlaEtAl_2018_Transliteration_Better_than_Translation_Answering_Code-mixed_Questions_over_Knowledge_Base,BanerjeeNaskarEtAl_First_Cross-Script_Code-Mixed_Question_Answering_Corpus,ChanduLoginovaEtAl_2018_Code-Mixed_Question_Answering_Challenge_Crowd-sourcing_Data_and_Techniques}, Telugu and Tamil~\cite{ChanduLoginovaEtAl_2018_Code-Mixed_Question_Answering_Challenge_Crowd-sourcing_Data_and_Techniques}.
\section{QA/RC ``skills''}
\label{sec:reasoning}

\subsection{Existing taxonomies}
\label{sec:reasoning-current}

We discussed above how different QA and RC datasets may be based on different understandings of ``format'' (\cref{sec:format-task}) and ``domain'' (\cref{sec:domains}), but there is even less agreement on what capabilities the QA/RC resources are intended to encapsulate. While nearly every paper presenting a RC or QA dataset also presents some exploratory data analysis of a small sample of their data, the categories they employ vary too much to enable direct comparisons between resources.

Traditionally, AI literature focused on types of ``reasoning'', which in philosophy and logic is defined as ``any process of drawing a conclusion from a set of premises may be called a process of reasoning''~\cite{Blackburn_2008_Reasoning}. Note that this is similar to the definition of ``inference'': ``the process of moving from (possibly provisional) acceptance of some propositions, to acceptance of others''~\cite{Blackburn_2008_Inference}. But the current QA/RC literature often discusses ``skills'' rather than ``reasoning types'', as many phenomena are defined on another level of analysis (e.g. linguistic phenomena such as coreference). 

To date, two taxononomies for the QA/RC ``skills'' have been proposed in the NLP literature:
\begin{itemize}
    \item \citet{SugawaraAizawa_2016_Analysis_of_Prerequisite_Skills_for_Reading_Comprehension,SugawaraKidoEtAl_2017_Evaluation_Metrics_for_Machine_Reading_Comprehension_Prerequisite_Skills_and_Readability} distinguish between object tracking skills, mathematical reasoning, logical reasoning, analogy, causal and spatiotemporal relations, ellipsis, bridging, elaboration, meta-knowledge, schematic clause relation, punctuation. 
    \item \citet{SchlegelValentinoEtAl_2020_Framework_for_Evaluation_of_Machine_Reading_Comprehension_Gold_Standards} distinguish between operational (bridge, constraint, comparison, intersection), arithmetic (subtraction, addition, ordering, counting, other), and linguistic (negation, quantifiers, conditional monotonicity, con/dis-junction) meta-categories, as opposed to temporal, spatial, causal reasoning, reasoning ``by exclusion'' and ``retrieval''. They further describe questions in terms of knowledge (factual/intuitive) and linguistic complexity (lexical and syntactic variety, lexical and syntactic ambiguity). 
\end{itemize}

A problem with any taxonomy is that using it to characterize new and existing resources involves expensive fine-grained expert annotation. A frequently used workaround is a kind of keyword analysis by the initial words in the question (since for English that would mean \textit{what}, \textit{where}, \textit{when} and other question words). This was done e.g. in \cite[e.g.][]{BajajCamposEtAl_2016_MS_MARCO_Human_Generated_MAchine_Reading_COmprehension_Dataset,KociskySchwarzEtAl_2018_NarrativeQA_Reading_Comprehension_Challenge,OstermannRothEtAl_2018_SemEval-2018_Task_11_Machine_Comprehension_Using_Commonsense_Knowledge}, and \citet{DzendzikVogelEtAl_2021_English_Machine_Reading_Comprehension_Datasets_Survey} perform such an analysis across 60 datasets, showing that $\approx$22.5\% of all questions are ``what'' questions. However, it is a characterization of the \emph{answers} to the questions, rather than \emph{the process} used to answer the question. It is also a very crude heuristic for the semantic type: for instance, ``what'' questions could target not only entities, but also properties (\textit{what color?}), locations (\textit{what country?}), temporal information (\textit{what day?}), etc.

\subsection{Proposed taxonomy}
\label{sec:reasoning-taxonomy}

Building on prior work~\cite{SugawaraAizawa_2016_Analysis_of_Prerequisite_Skills_for_Reading_Comprehension,SugawaraKidoEtAl_2017_Evaluation_Metrics_for_Machine_Reading_Comprehension_Prerequisite_Skills_and_Readability,SchlegelValentinoEtAl_2020_Framework_for_Evaluation_of_Machine_Reading_Comprehension_Gold_Standards}, \anna{we propose an alternative taxonomy. It accounts for a wider range of of QA/RC ``skills'' (e.g. including the multi-step reasoning), and we believe it offers a more systematic grouping of ``skills'' along the following top-level dimensions}: 

\begin{itemize*}
\item \textbf{Inference} (\cref{sec:reasoning-inference}): ``the process of moving from (possibly provisional) acceptance of some propositions, to acceptance of others''~\cite{Blackburn_2008_Inference}.
\item \textbf{Retrieval} (\cref{sec:reasoning-retrieval}): %
knowing where to look for the relevant information.
\item \textbf{Input interpretation \& manipulation} (\cref{sec:reasoning-interpreting}): correctly understanding the meaning of all the signs in the input, both linguistic and numeric, and performing any operations on them that are defined by the given language/mathematical system (identifying coreferents, summing up etc.).
\item \textbf{World modeling} (\cref{sec:reasoning-world}): constructing a valid representation of the spatiotemporal and social aspects of the world described in the text, as well as positioning the text itself with respect to the reader and other texts.
\item \textbf{Multi-step} (\cref{sec:multi-step}): performing chains of actions on any of the above dimensions.
\end{itemize*}

A key feature of our taxonomy is that these dimensions are \textit{orthogonal}: the same question can be described in terms of their linguistic form, the kind of inference required to arrive at the answer, retrievability of the evidence, compositional complexity, and the level of world modeling (from generic open-domain questions to questions about character relations in specific books). In a given question, some of them may be more prominent/challenging than others. %

\begin{figure}
    \centering
    \includegraphics[width=\linewidth]{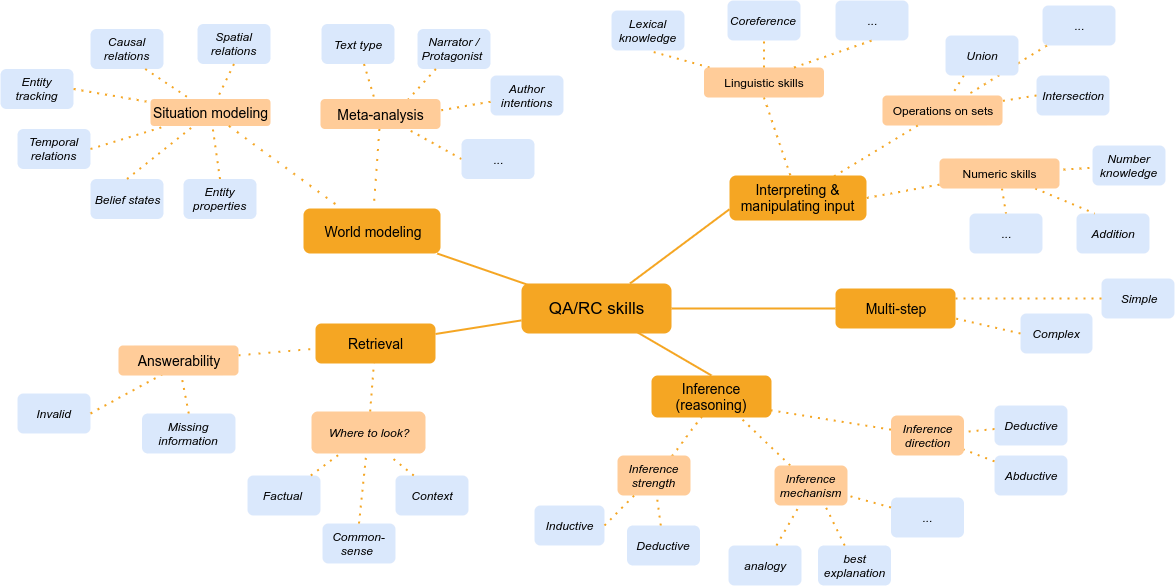}
    \caption{Proposed classification of machine reading comprehension skills}
    \label{fig:reasoning}
\end{figure}

Our proposal is shown in \autoref{fig:reasoning} and discussed in more detail below. 

\subsubsection{Inference type.}
\label{sec:reasoning-inference}
Fundamentally, QA can be conceptualized as the classification of the relation between the premise (context+question) and the conclusion (a candidate answer)~\cite{SachanDubeyEtAl_2015_Learning_Answer-Entailing_Structures_for_Machine_Comprehension}. Then, the type of reasoning performed by the system can be categorized in the terms developed in logic and philosophy\footnote{Note that logical reasoning is only a subset of  human reasoning: human decisions are not necessarily rational, they may be based on biases, heuristics, or fall pray to different logical fallacies. It is not clear to what extent the human reasoning ``shortcuts'' should be replicated in machine RC systems, especially given that they develop their own biases and heuristics (see \cref{sec:discussion-required-skills}).}. Among the criteria developed for describing different types of reasoning is the \textit{direction of reasoning}: deductive (from premise to conclusion) and abductive (from conclusion to the premise that would best justify the conclusion)~\cite{Douven_2017_Abduction}. Another key criterion is \textit{the degree to which the premise supports the conclusion}: in deductive reasoning, the hypothesis is strictly entailed by the premise, and in inductive reasoning, the support is weaker~\cite{Hawthorne_2021_Inductive_Logic}. Finally, reasoning could be analysed with respect to the \textit{kind of support for the conclusion}, including analogical reasoning~\cite{Bartha_2019_Analogy_and_Analogical_Reasoning}, defeasible reasoning (``what normally\footnote{There is debate on whether probabilistic inference belongs to inductive or deductive logic~\cite{DemeyKooiEtAl_2019_Logic_and_Probability}, which we leave to the philosophers. But SEP defines defeasible reasoning as non-deductive reasoning based on ``what normally happens''~\cite{Koons_2017_Defeasible_Reasoning}, which seems to presuppose the notion of probability.} happens'',~\cite{Koons_2017_Defeasible_Reasoning}), and ``best explanation''~\cite{Douven_2017_Abduction}. 

While the above criteria are among the most fundamental and well-recognized to describe human reasoning, none of them is actively used to study machine reasoning, at least in the current QA/RC literature. Even though deductive reasoning is both fundamental and the most clearly mappable to what we could expect from machine reasoning, to the best of our knowledge so far there is only one dataset for that: LogiQA~\cite{LiuCLHWZ20}, a collection of multi-choice questions from civil servant exam materials.

To further complicate the matter, sometimes the above-mentioned terms are even used differently. For instance, ReClor~\cite{YuJiangEtAl_2019_ReClor_Reading_Comprehension_Dataset_Requiring_Logical_Reasoning} is presented as a resource targeting logical reasoning, but it is based on GMAT/LSAT teaching materials, and much of it actually targets meta-analysis of the logical structure rather than logical reasoning itself (e.g. identifying claims and conclusions in the provided text). CLUTTR~\cite{SinhaSodhaniEtAl_2019_CLUTRR_Diagnostic_Benchmark_for_Inductive_Reasoning_from_Text} is an inductive reasoning benchmark for kinship relations, but the term ``inductive'' is used in the sense of ``inducing rules'' (similar to the above definition of ``inference'') rather than as ``non-deductive'' (i.e. offering only partial support for the conclusion).

A kind of non-deductive reasoning that historically received a lot of attention in the AI literature is defeasible reasoning~\cite{McCarthyHayes_1969_Some_Philosophical_Problems_From_Standpoint_of_Artificial_Intelligence,ChesnevarMaguitmanEtAl_2000_Logical_models_of_argument}, which is now making a comeback in NLI~\cite{RudingerShwartzEtAl_2020_Thinking_Like_Skeptic_Defeasible_Inference_in_Natural_Language} (formulated as the task of re-evaluating the strength of the conclusion in the light of an additional premise strengthening/weakening the evidence offered by the original premise). There is also ART~\cite{BhagavatulaBrasEtAl_2019_Abductive_Commonsense_Reasoning}, an abductive reasoning challenge where the system needs to come up with a hypothesis that better complements incomplete observations.

\subsubsection{Retrieval.}
\label{sec:reasoning-retrieval}
It could be argued that information retrieval happens \textit{before} inference: to evaluate a premise and a conclusion, we first have to have them. But inference can also be viewed as the \textit{ranking mechanism} of retrieval: NLP systems consider the answer options so as to choose the one offering the strongest support for the conclusion. This is how the current systems approach close-world reading comprehension tests like RACE~\cite{LaiXieEtAl_2017_RACE_Large-scale_ReAding_Comprehension_Dataset_From_Examinations} or SQuAD~\cite{RajpurkarZhangEtAl_2016_SQuAD_100000+_Questions_for_Machine_Comprehension_of_Text}. In the open-world setting, instead of a specific text, we have a much broader set of options (a corpus of snippets, a knowledge base, knowledge encoded by a language model etc.). However, fundamentally the task is still to find the best answer out of the available knowledge. We are considering two sub-dimensions of the retrieval problem: determining whether an answer exists, and where to look for it. 

\textbf{Answerability.} SQuAD 2.0~\cite{RajpurkarJiaEtAl_2018_Know_What_You_Dont_Know_Unanswerable_Questions_for_SQuAD} popularized the distinction between questions that are answerable with the given context and those that are not. Arguably, the distinction is actually not binary, and at least two resources argue for a 3-point uncertainty scale. ReCO~\cite{WangYaoEtAl_2020_ReCO_Large_Scale_Chinese_Reading_Comprehension_Dataset_on_Opinion} offers boolean questions with ``yes'', ``no'' and ``maybe'' answer options. QuAIL~\cite{RogersKovalevaEtAl_2020_Getting_Closer_to_AI_Complete_Question_Answering_Set_of_Prerequisite_Real_Tasks} distinguishes between full certainty (answerable with a given context), partial certainty (a confident guess can be made with a given context + some external common knowledge), and full uncertainty (no confident guess can be made even with external common knowledge). A more general definition of the unanswerable questions would be this: the questions that cannot be answered \textit{given all the information that the reader has access to}.

This is different from \textit{invalid questions}: the questions that a human would reject rather than attempt to answer. \autoref{tab:invalid-questions} shows examples of different kinds of violations: the answers that are impossible to retrieve, loaded questions, ill-formed questions, rhetorical questions, ``useless'' questions, and others.

\textbf{Where to look for the target knowledge?} The classical RC case in resources like SQuAD is a single \textit{context} that is the only possible source of information: in this case, the retrieval problem is reduced to finding the relevant span. When the knowledge is not provided, the system needs to know where to find it,\footnote{For human readers, \citet{McNamaraMagliano_2009_Chapter_9_Toward_Comprehensive_Model_of_Comprehension} similarly distinguish between \textit{bridging} (linking new information---in this case, from the question---to previous context) and \textit{elaboration} (linking information to some external information).} and in this case it may be useful to know whether it is \textit{factual} (e.g. ``Dante was born in Florence'') or \textit{world knowledge} (e.g. ``bananas are yellow'').\footnote{\citet{SchlegelNenadicEtAl_2020_Beyond_Leaderboards_survey_of_methods_for_revealing_weaknesses_in_Natural_Language_Inference_data_and_models} distinguish between ``factual'' and ``intuitive'' knowledge. The latter is defined as that ``which is challenging to express as a set of facts, such as the knowledge that a parenthetic numerical expression next to a person’s name in a biography usually denotes [their] life span''.} This is the core distinction between the subfields of open-domain QA and commonsense reasoning, respectively. Note that in both of those cases, the source of knowledge is external to the question and must be retrieved from somewhere (Web snippets, knowledge bases, model weights, etc.). The difference is in the \textit{human} competence: an average human speaker is not expected to have all the factual knowledge, but is expected to have a store of the world knowledge (even though the specific subset of that knowledge is culture- and age-dependent). 

\begin{table}
    \centering
    \begin{tabular}{p{8.6cm} p{5.8cm}}
    \toprule
    \textit{Example} & \textit{Problem} \\
    \midrule
    Have you stopped beating your wife? & invalid premise (that the wife is beaten) \\
    What is the meaning of life, the universe, and everything?~\cite{Adams_2009_hitchhikers_guide_to_galaxy} & not specific enough \\    
    At what age can I get a driving license? & missing information (in what country?) \\
    Can quantum mechanics and relativity be linked together? & information not yet discovered \\
    What was the cause of the US civil war?~\cite{Boyd-Graber_2019_What_Question_Answering_can_Learn_from_Trivia_Nerds} & no consensus on the answer \\ 
    Who can figure out the true meaning of `covfefe'? & %
    uninterpretable due to language errors \\
    Do colorless ideas sleep furiously? & syntactically well-formed but uninterpretable \\
    What is the sum of angles in a triangle with sides 1, 1, and 10~cm?\footnotemark
    & such a triangle cannot exist\\
    What have the Romans ever done for us? \cite{ChapmanCleeseEtAl_1999_Life_of_Brian} & rhetorical question \\
    What is the airspeed velocity of a swallow carrying a coconut? \cite{WhiteChapmanEtAl_2001_Monty_Python_and_Holy_Grail} & the answer would not be useful\footnotemark
    to know\\
    \bottomrule
    \end{tabular}
    \caption{Types of invalid questions}
    \label{tab:invalid-questions}
\end{table}

\footnotetext[12]{\url{https://philosophy.stackexchange.com/questions/37311/are-all-answers-to-a-contradictory-question-correct-or-are-all-wrong-or-is-it}}
\footnotetext[13]{The practical utility of questions is hard to estimate objectively, given that human interests vary a lot.  \citet{HorbachAldabeEtAl_2020_Linguistic_Appropriateness_and_Pedagogic_Usefulness_of_Reading_Comprehension_Questions} annotate questions for centrality to the given topic, and whether a teacher would be likely to use that question with human students, but the human agreement on their sample is fairly low. The agreement is likely even less for the more niche, specialist questions: the low agreement on acceptance recommendations in peer review~\cite{Price_2014_NIPS_experiment} is likely partly due to the fact that different groups of researchers simply do not find each other's research questions equally exciting.}

Many resources for the former were discussed in \cref{sec:domains}. Commonsense reasoning resources deserve a separate survey, but overall, most levels of description discussed in this paper also apply to them. They have the analog of open-world factoid QA (e.g. CommonsenseQA~\cite{TalmorHerzigEtAl_2019_CommonsenseQA_Question_Answering_Challenge_Targeting_Commonsense_Knowledge}, where the task is to answer a multi-choice question without any given context), but more resources are described as ``reading comprehension'', with multi-choice~\cite{OstermannRothEtAl_2018_SemEval-2018_Task_11_Machine_Comprehension_Using_Commonsense_Knowledge,HuangLeBrasEtAl_2019_Cosmos_QA_Machine_Reading_Comprehension_with_Contextual_Commonsense_Reasoning} or cloze-style~\cite{ZhangLiuEtAl_2018_ReCoRD_Bridging_Gap_between_Human_and_Machine_Commonsense_Reading_Comprehension} questions asked in the context of some provided text. %
Similarly to ``domains'' in open-world QA (see \cref{sec:domains}), there are specialist resources targeting specific types of world knowledge (see \cref{sec:reasoning-world}).

\subsubsection{Interpreting \& manipulating input.}
\label{sec:reasoning-interpreting}
This dimension necessarily applies to any question: both humans and machines \textit{should} have the knowledge of the meaning of the individual constituent elements of the input (words, numbers), and have the ability to perform operations on them that are defined by the language/shared mathematical system (rather than given in the input).\footnote{The current NLP systems can perform well on QA/RC benchmarks even when they are transformed to become uninterpretable to humans~\cite{SugawaraStenetorpEtAl_2020_Assessing_Benchmarking_Capacity_of_Machine_Reading_Comprehension_Datasets}. It is an open question whether we should strive for systems to reject inputs that a human would reject, and on the same grounds.} It includes the following subcategories:

\begin{itemize*}
\item \textbf{Linguistic skills.} SQuAD \cite{RajpurkarZhangEtAl_2016_SQuAD_100000+_Questions_for_Machine_Comprehension_of_Text}, one of the first major RC resources, predominantly targeted argument extraction and event paraphrase detection. Curently many resources focus on coreference resolution (e.g. Quoref~\cite{DasigiLiuEtAl_2019_Quoref_Reading_Comprehension_Dataset_with_Questions_Requiring_Coreferential_Reasoning}, part of DROP~\cite{DuaWangEtAl_2019_DROP_Reading_Comprehension_Benchmark_Requiring_Discrete_Reasoning_Over_Paragraphs}). Among the reasoning types proposed in~\cite{SugawaraAizawa_2016_Analysis_of_Prerequisite_Skills_for_Reading_Comprehension,SugawaraKidoEtAl_2017_Evaluation_Metrics_for_Machine_Reading_Comprehension_Prerequisite_Skills_and_Readability}, ``linguistic skills'' also include ellipsis, schematic clause relations, punctuation. The list is not exhaustive: arguably, any questions formulated in a natural language depend on a large number of linguistic categories (e.g. reasoning about temporal relations must involve knowledge of verb tense), and even the questions targeting a single phenomenon as it is defined in linguistics (e.g. coreference resolution) do also require other linguistic skills (e.g. knowledge of parts of speech). Thus, any analysis based on linguistic skills should allow the same question to belong to several categories, and it is not clear whether we can reliably determine which of them are more ``central''. \\
Questions (and answers/contexts) could also be characterized in terms of ``ease of processing''~\cite{McNamaraMagliano_2009_Chapter_9_Toward_Comprehensive_Model_of_Comprehension}, which is related to the set of linguistic phenomena involved in its surface form. But it does not mean the same thing for humans and machines: the latter have a larger vocabulary, do not get tired in the same way, etc.
\item \textbf{Numeric skills.} In addition to the linguistic knowledge required for interpreting numeral expressions, an increasing number of datasets is testing NLP systems' abilities of answering questions that require mathematical operations over the information in the question and the input context. DROP~\cite{DuaWangEtAl_2019_DROP_Reading_Comprehension_Benchmark_Requiring_Discrete_Reasoning_Over_Paragraphs} involves numerical reasoning over multiple paragraphs of Wikipedia texts. \citet{MishraMitraEtAl_2020_Towards_Question_Format_Independent_Numerical_Reasoning_Set_of_Prerequisite_Tasks} contribute a collection of small-scale numerical reasoning datasets including extractive, freeform, and multi-choice questions, some of them requiring retrieval of external world knowledge. There is also a number of resources targeting school algebra word problems~\cite{KushmanArtziEtAl_2014_Learning_to_Automatically_Solve_Algebra_Word_Problems,ShiWangEtAl_2015_Automatically_Solving_Number_Word_Problems_by_Semantic_Parsing_and_Reasoning,UpadhyayChang_2017_Annotating_Derivations_New_Evaluation_Strategy_and_Dataset_for_Algebra_Word_Problems,MiaoLiangEtAl_2020_Diverse_Corpus_for_Evaluating_and_Developing_English_Math_Word_Problem_Solvers} and multimodal counting benchmarks~\cite{ChattopadhyayVedantamEtAl_2017_Counting_Everyday_Objects_in_Everyday_Scenes,AcharyaKafleEtAl_2019_TallyQA_Answering_Complex_Counting_Questions}. FinQA~\cite{ChenChenEtAl_2021_FinQA_Dataset_of_Numerical_Reasoning_over_Financial_Data} and Tat-QA~\cite{ZhuLeiEtAl_2021_TAT-QA} present the challenge of numerical reasoning over financial documents (also containing tables).
\item \textbf{Operations on sets.} This category targets such operations as union, intersection, ordering, and determining subset/superset relations which going beyond the lexical knowledge subsumed by the hypernymy/hyponymy relations. The original bAbI~\cite{WestonBordesEtAl_2015_Towards_AIcomplete_question_answering_A_set_of_prerequisite_toy_tasks} included ``lists/sets'' questions such as \textit{Daniel picks up the football. Daniel drops the newspaper. Daniel picks up the milk. John took the apple. What is Daniel holding? (milk, football)}. Among the categories proposed by \citet{SchlegelNenadicEtAl_2020_Beyond_Leaderboards_survey_of_methods_for_revealing_weaknesses_in_Natural_Language_Inference_data_and_models}, the ``constraint'' skill is fundamentally the ability to pick a subset the members which satisfy an extra criterion.
\end{itemize*}

Some linguistic phenomena highly correlate with certain reasoning operations, but overall these two dimensions are still orthogonal. A prime example is comparison:\footnote{So far comparison is directly targeted in QuaRel~\cite{TafjordClarkEtAl_2019_QuaRel_Dataset_and_Models_for_Answering_Questions_about_Qualitative_Relationships}, and also present in parts of other resources~\cite{SchlegelValentinoEtAl_2020_Framework_for_Evaluation_of_Machine_Reading_Comprehension_Gold_Standards,DuaWangEtAl_2019_DROP_Reading_Comprehension_Benchmark_Requiring_Discrete_Reasoning_Over_Paragraphs}.} it is often expressed with comparative degrees of adjectives (in the question or context) and so requires interpretation of those linguistic signs. At the same time, unless the answer is directly stated in the text, it also requires a deductive inference operation. For example: \textit{John wears white, Mary wears black. Who wears darker clothes?}
 
\subsubsection{World modeling.}
\label{sec:reasoning-world}
One of major psychological theories of human RC is based on mental simulation: when we read, we create a model of the described world, which requires that we ``instantiate'' different objects and entities, track their locations, and ingest and infer the temporal and causal relations between events~\cite{vanderMeerBeyerEtAl_2002_Temporal_order_relations_in_language_comprehension,Zwaan_2016_Situation_models_mental_simulations_and_abstract_concepts_in_discourse_comprehension}. Situation modeling has been proposed as one of the levels of representation in discourse comprehension~\cite{DijkKintsch_1983_Strategies_of_discourse_comprehension}, and it is the basis for the recent ``templates of understanding''~\cite{DunietzBurnhamEtAl_2020_To_Test_Machine_Comprehension_Start_by_Defining_Comprehension} that include spatial, temporal, causal and motivational elements. We further add %
the category of belief states~\cite{RogersKovalevaEtAl_2020_Getting_Closer_to_AI_Complete_Question_Answering_Set_of_Prerequisite_Real_Tasks}, since human readers keep track not only of spatiotemporal and causal relations in a narrative, but also the who-knows-what information.

A challenge for psychological research is that different kinds of texts have a different mixture of prominent elements (temporal structure for narratives, referential elements in expository texts etc.), and the current competing models were developed on the basis of different kinds of evidence, which makes them hard to reconcile~\cite{McNamaraMagliano_2009_Chapter_9_Toward_Comprehensive_Model_of_Comprehension}. This is also the case for machine RC, and partly explains the lack of agreement about classification of ``types of reasoning'' across the literature. Based on our classification, the following resources explicitly target a specific aspect of situation modeling, in either RC (i.e. ``all the necessary information in the text'') or commonsense reasoning (i.e. ``text needs to be combined with extra world knowledge'') settings:\footnote{This list does not contain all possible types of information in world modeling category, and includes only the resources that specifically target a given type of information, or have a part targeting that type of information that can be separated based on the provided annotation.}

\begin{itemize*}
\item spatial reasoning: bAbI~\cite{WestonBordesEtAl_2015_Towards_AIcomplete_question_answering_A_set_of_prerequisite_toy_tasks}, SpartQA~\cite{mirzaee-etal-2021-spartqa}, many VQA datasets \cite[e.g.][see \cref{sec:media}]{JangSongEtAl_2017_TGIF-QA_Toward_Spatio-Temporal_Reasoning_in_Visual_Question_Answering};
\item temporal reasoning: event order (QuAIL~\cite{RogersKovalevaEtAl_2020_Getting_Closer_to_AI_Complete_Question_Answering_Set_of_Prerequisite_Real_Tasks}, TORQUE~\cite{NingWuEtAl_2020_TORQUE_Reading_Comprehension_Dataset_of_Temporal_Ordering_Questions}), event attribution to time (TEQUILA~\cite{JiaAbujabalEtAl_2018_TEQUILA_Temporal_Question_Answering_over_Knowledge_Bases}, TempQuestions~\cite{JiaAbujabalEtAl_2018_TempQuestions_Benchmark_for_Temporal_Question_Answering}, script knowledge (MCScript~\cite{OstermannRothEtAl_2018_SemEval-2018_Task_11_Machine_Comprehension_Using_Commonsense_Knowledge}), event duration (MCTACO~\cite{ZhouKhashabiEtAl_2019_Going_on_vacation_takes_longer_than_Going_for_walk_Study_of_Temporal_Commonsense_Understanding}, QuAIL~\cite{RogersKovalevaEtAl_2020_Getting_Closer_to_AI_Complete_Question_Answering_Set_of_Prerequisite_Real_Tasks}), temporal commonsense knowledge (MCTACO~\cite{ZhouKhashabiEtAl_2019_Going_on_vacation_takes_longer_than_Going_for_walk_Study_of_Temporal_Commonsense_Understanding}, TIMEDIAL~\cite{qin2021timedial}), \final{factoid/news questions with answers where the correct answers change with time (ArchivalQA ~\cite{WangJatowtEtAl_2022_ArchivalQA_Large-scale_Benchmark_Dataset_for_Open_Domain_Question_Answering_over_Historical_News_Collections}, SituatedQA~\cite{ZhangChoi_2021_SituatedQA_Incorporating_Extra-Linguistic_Contexts_into_QA})}, temporal reasoning in multimodal setting~\cite{JangSongEtAl_2017_TGIF-QA_Toward_Spatio-Temporal_Reasoning_in_Visual_Question_Answering,FayekJohnson_2020_Temporal_Reasoning_via_Audio_Question_Answering};
\item belief states: Event2Mind~\cite{RashkinSapEtAl_2018_Event2Mind_Commonsense_Inference_on_Events_Intents_and_Reactions}, QuAIL~\cite{RogersKovalevaEtAl_2020_Getting_Closer_to_AI_Complete_Question_Answering_Set_of_Prerequisite_Real_Tasks};
\item causal relations: ROPES~\cite{LinTafjordEtAl_2019_Reasoning_Over_Paragraph_Effects_in_Situations}, QuAIL~\cite{RogersKovalevaEtAl_2020_Getting_Closer_to_AI_Complete_Question_Answering_Set_of_Prerequisite_Real_Tasks}, QuaRTz~\cite{tafjord-etal-2019-quartz}, ESTER~\cite{HanHsuEtAl_2021_ESTER_Machine_Reading_Comprehension_Dataset_for_Reasoning_about_Event_Semantic_Relations};
\item \final{other relations between events: subevents, conditionals, counterfactuals etc.~\cite{HanHsuEtAl_2021_ESTER_Machine_Reading_Comprehension_Dataset_for_Reasoning_about_Event_Semantic_Relations};}
\item entity properties and relations:\footnote{QA models are even used directly for relation extraction~\cite{LevySeoEtAl_2017_Zero-Shot_Relation_Extraction_via_Reading_Comprehension,abdou-etal-2019-x,LiYinEtAl_2019_Entity-Relation_Extraction_as_Multi-Turn_Question_Answering}.} social interactions (SocialIQa~\cite{SapRashkinEtAl_2019_Social_IQa_Commonsense_Reasoning_about_Social_Interactions}), properties of characters (QuAIL ~\cite{RogersKovalevaEtAl_2020_Getting_Closer_to_AI_Complete_Question_Answering_Set_of_Prerequisite_Real_Tasks}), physical properties (PIQA~\cite{BiskZellersEtAl_2020_PIQA_Reasoning_about_Physical_Commonsense_in_Natural_Language}, QuaRel~\cite{TafjordClarkEtAl_2019_QuaRel_Dataset_and_Models_for_Answering_Questions_about_Qualitative_Relationships}), numerical properties (NumberSense~\cite{LinLeeEtAl_2020_Birds_have_four_legs_NumerSense_Probing_Numerical_Commonsense_Knowledge_of_Pre-Trained_Language_Models});
\item tracking entities: across locations (bAbI~\cite{WestonBordesEtAl_2015_Towards_AIcomplete_question_answering_A_set_of_prerequisite_toy_tasks}), in coreference chains (Quoref~\cite{DasigiLiuEtAl_2019_Quoref_Reading_Comprehension_Dataset_with_Questions_Requiring_Coreferential_Reasoning}, resources in the Winograd Schema Challenge family~\cite{LevesqueDavisEtAl_2012_Winograd_Schema_Challenge,SakaguchiBrasEtAl_2019_WINOGRANDE_Adversarial_Winograd_Schema_Challenge_at_Scale}). Arguably the cloze-style resources based on named entities also fall into this category (CBT~\cite{HillBordesEtAl_2015_Goldilocks_Principle_Reading_Childrens_Books_with_Explicit_Memory_Representations}, CNN/DailyMail~\cite{HermannKociskyEtAl_2015_Teaching_Machines_to_Read_and_Comprehend}, WhoDidWhat~\cite{OnishiWangEtAl_2016_Who_did_What_Large-Scale_Person-Centered_Cloze_Dataset}), but they do not guarantee that the masked entity is in some complex relation with its context.
\end{itemize*}

The text + alternative endings format used in several commonsense datasets like SWAG (see \cref{sec:format-ending}) has the implicit question ``What happened next?''.
These resources cross-cut causality and temporality: much of such data seems to target causal relations (specifically, the knowledge of possible effects of interactions between characters and entities), but also script knowledge, and the format clearly presupposes the knowledge of the temporal before/after relation.

A separate aspect of world modeling is the meta-analysis skills: the ability of the reader to identify the likely time, place and intent of its writer, the narrator, the protagonist/antagonist, identifying stylistic features and other categories. These skills are considered as a separate category by \citet{SugawaraKidoEtAl_2017_Evaluation_Metrics_for_Machine_Reading_Comprehension_Prerequisite_Skills_and_Readability}, and are an important target of the field of literary studies, but so far they have not been systematically targeted in machine RC. That being said, some existing resources include questions formulated to include words like ``author'' and ``narrator''~\cite{RogersKovalevaEtAl_2020_Getting_Closer_to_AI_Complete_Question_Answering_Set_of_Prerequisite_Real_Tasks}. They are also a part of some resources that were based on existing pedagogical resources, such as some of ReClor~\cite{YuJiangEtAl_2019_ReClor_Reading_Comprehension_Dataset_Requiring_Logical_Reasoning} questions
that focus on identifying claims and conclusions in the provided text.

\subsubsection{Multi-step reasoning.} 
\label{sec:multi-step}

Answering a question may require one or several pieces of information. In the recent years a lot of attention was drawn to what could be called multi-step information retrieval, with resources focusing on ``simple'' and ``complex'' questions:
\begin{itemize}
    \item ``Simple'' questions have been defined as such that ``refer to a single fact of the KB''~\cite{BordesUsunierEtAl_2015_Large-scale_Simple_Question_Answering_with_Memory_Networks}. In an RC context, this corresponds to the setting where all the necessary evidence is contained in the same place in the text. 
    \item The complex questions, accordingly, are the questions that rely on several facts~\cite{TalmorBerant_2018_Web_as_Knowledge-Base_for_Answering_Complex_Questions}. In an RC setting, this corresponds to the so-called multi-hop datasets that necessitate the combination of information across sentences~\cite{KhashabiChaturvediEtAl_2018_Looking_Beyond_Surface_Challenge_Set_for_Reading_Comprehension_over_Multiple_Sentences}, paragraphs~\cite{DuaWangEtAl_2019_DROP_Reading_Comprehension_Benchmark_Requiring_Discrete_Reasoning_Over_Paragraphs}, and documents~\cite{YangQiEtAl_2018_HotpotQA_Dataset_for_Diverse_Explainable_Multi-hop_Question_Answering}. It also by definition includes questions that require a combination of context and world knowledge \cite[e.g.][]{RogersKovalevaEtAl_2020_Getting_Closer_to_AI_Complete_Question_Answering_Set_of_Prerequisite_Real_Tasks}.
\end{itemize}

That being said, the ``multi-step'' skill seems broader than simply combining several facts, and could also be taken as combining the ``skills'' from different dimensions of our taxonomy. In particular, any question is a linguistically complex expression, and so the success of any retrieval steps depends on understanding the various parts of the question and how they are combined. Consider the question ``Who played Sherlock Holmes, starred in Avengers and was born in London?": it requires several retrieval steps, but before we can even get to that, we have to perform some kind of semantic parsing to even interpret the question.

\section{Discussion}
\label{sec:discussion}

This section concludes the paper with broader discussion of reasoning skills: %
the types of ``skills'' that are minimally required for our systems to solve QA/RC benchmarks (\cref{sec:discussion-required-skills}) vs the ones that a human would use (\cref{sec:analysis}). We then proceed to highlighting the gaps in the current research, specifically the kinds of datasets that have not been made yet (\cref{sec:missing-data}).%

\subsection{What reasoning skills are actually required?}
\label{sec:discussion-required-skills}

A key assumption in the current analyses of QA/RC data in terms of the capabilities that they target (including our own taxonomy in \cref{sec:reasoning-taxonomy}) is that humans and models would use the same ``skills'' to answer a given question. But that is not necessarily true. DL models search for patterns in the training data, and they may and do find various shortcuts that happen to also predict the correct answer \cite[][inter alia]{GururanganSwayamdiptaEtAl_2018_Annotation_Artifacts_in_Natural_Language_Inference_Data,JiaLiang_2017_Adversarial_Examples_for_Evaluating_Reading_Comprehension_Systems,GevaGoldbergEtAl_2019_Are_We_Modeling_Task_or_Annotator_Investigation_of_Annotator_Bias_in_Natural_Language_Understanding_Datasetsa,SugawaraStenetorpEtAl_2020_Assessing_Benchmarking_Capacity_of_Machine_Reading_Comprehension_Datasets}%
. An individual question may well target e.g. coreference, but if it contains a word that is consistently associated with the first answer option in a multi-choice dataset, the model could potentially answer it without knowing anything about coreference. What is worse, \textit{how} a given question is answered could change with a different split of the same dataset, a model with a different inductive bias, or, the most frustratingly, even a different run of the same model~\cite{McCoyMinEtAl_2019_BERTs_of_feather_do_not_generalize_together_Large_variability_in_generalization_across_models_with_similar_test_set_performance}.

This means that there is a discrepancy between the reasoning skills that a question seems to target, and the skills that are minimally required to ``solve'' a particular dataset. In the context of the traditional machine learning workflow with training and testing, \textit{we need to reconsider the idea that whether or not a given reasoning skill is ``required'' is a characteristic of a given question. It is rather a characteristic of the combination of that question and the entire dataset}.

The same limitation applies to the few-shot- or in-context-learning paradigm based on large language models~\cite{BrownMannEtAl_2020_Language_Models_are_Few-Shot_Learners}, where only a few samples of the target task are presented as examples and no gradient updates are performed. Conceptually, such models still encapsulate the patterns observed in their training data, and so may still be choosing the correct answer option e.g. because there were more training examples with the correct answer listed first%
~\cite{ZhaoWallaceEtAl_2021_Calibrate_Before_Use_Improving_Few-Shot_Performance_of_Language_Models}%
. The difference is only that it is much harder to perform the training data analysis and find any such superficial hints.

How can we ever tell that the model is producing the correct answer for the right reasons? There is now enough work in this area to deserve its own survey, but the main directions are roughly as follows:

\begin{itemize*}
    \item construction of \textit{diagnostic tests}: adversarial tests \cite[e.g.][]{JiaLiang_2017_Adversarial_Examples_for_Evaluating_Reading_Comprehension_Systems,WallaceBoyd-Graber_2018_Trick_Me_If_You_Can_Adversarial_Writing_of_Trivia_Challenge_Questions}, probes for specific linguistic or reasoning skills \cite[e.g.][]{RibeiroWuEtAl_2020_Beyond_Accuracy_Behavioral_Testing_of_NLP_Models_with_CheckList,SinhaSodhaniEtAl_2019_CLUTRR_Diagnostic_Benchmark_for_Inductive_Reasoning_from_Text,JimenezRussakovskyEtAl_2022_CARETS_Consistency_And_Robustness_Evaluative_Test_Suite_for_VQA}, minimal pair evaluation around the model's decision boundary \cite[e.g.][]{KaushikHovyEtAl_2019_Learning_Difference_That_Makes_Difference_With_Counterfactually-Augmented_Data,GardnerArtziEtAl_2020_Evaluating_Models_Local_Decision_Boundaries_via_Contrast_Sets}.
    \item creating \textit{larger collections of generalization tests}, both out-of-domain (\cref{sec:domains}) and cross-lingual (\cref{sec:languages-multi})%
    . The assumption is that as their number grows the likelihood of the model solving them all with benchmark-specific heuristics decreases.
    \item work on \textit{controlling the signal in the training data}, on the assumption that if a deep learning model has a good opportunity to learn some phenomenon, it should do so (although that is not necessarily the case~\cite{GeigerCasesEtAl_2019_Posing_Fair_Generalization_Tasks_for_Natural_Language_Inference}). This direction includes all the work on resources focusing on specific reasoning or linguistic skills (\cite[e.g.][]{DasigiLiuEtAl_2019_Quoref_Reading_Comprehension_Dataset_with_Questions_Requiring_Coreferential_Reasoning}) and balanced sets of diverse skills \cite[e.g.][]{RogersKovalevaEtAl_2020_Getting_Closer_to_AI_Complete_Question_Answering_Set_of_Prerequisite_Real_Tasks,CaoShiEtAl_2022_KQA_Pro_Dataset_with_Explicit_Compositional_Programs_for_Complex_Question_Answering_over_Knowledge_Base}. \\ This direction also includes the methodology work on crafting the data to avoid reasoning shortcuts: e.g. using human evaluation to discard the questions that humans could answer without considering full context~\cite{PapernoKruszewskiEtAl_2016_LAMBADA_dataset_Word_prediction_requiring_broad_discourse_context}.
    \item \textit{interpretability work} on generating human-interpretable explanations for a given prediction, e.g. by context attribution \cite[e.g.][]{RibeiroSinghEtAl_2016_Why_should_i_trust_you_Explaining_predictions_of_any_classifier} or influential training examples \cite[e.g.][]{RajaniKrauseEtAl_2020_Explaining_and_Improving_Model_Behavior_with_k_Nearest_Neighbor_Representations}. However, the faithfulness of such explanations is itself an active area of research \cite[e.g.][]{PruthiGuptaEtAl_2019_Learning_to_Deceive_with_Attention-Based_Explanations,YinShiEtAl_2021_On_Faithfulness_Measurements_for_Model_Interpretations}. The degree to which humans can use explanations to evaluate the quality of the model also varies depending on the model quality and prior belief bias~\cite{Gonzales2021_belief-bias}. 
\end{itemize*}

\final{All of these research directions play an important role: our chances of success are higher with higher quality training data, we need better and more systematic tests, and we need faithful interpretability techniques to be able to tell what the model is doing. Eventually these different directions would ideally combine. \citet{ChoudhuryRogersEtAl_2022_Machine_Reading_Fast_and_Slow_When_Do_Models_Understand_Language} make the case for (a) defining what information the model should rely on to answer specific kinds of questions, (b) testing whether that is the case, both in and out of the training distribution.}

While we may never be able to say conclusively that a blackbox model relies on the same strategies as a human reader, we should (and, under the article 13 of the AI Act proposal, could soon be legally required to\footnote{\url{https://eur-lex.europa.eu/resource.html?uri=cellar:e0649735-a372-11eb-9585-01aa75ed71a1.0001.02/DOC_1&format=PDF}}%
) at least identify the cases in which they succeed and in which they fail, as it is prerequisite for safe deployment.

\subsection{Analysis of question types and reasoning skills}
\label{sec:analysis}

Section \ref{sec:discussion-required-skills} discussed the fundamental difficulties with identifying how a blackbox neural model was able to solve a QA/RC task. However, we also have trouble even identifying the processes a \textit{human} reader would use to answer a question. As discussed in \cref{sec:reasoning-current}, there are so far only two studies attempting cross-dataset analysis of reasoning skills according to a given skill taxonomy, and they both only target small samples (50-100 examples per resource). This is due to the fact that such analysis requires expensive expert annotation.  \citet{HorbachAldabeEtAl_2020_Linguistic_Appropriateness_and_Pedagogic_Usefulness_of_Reading_Comprehension_Questions} showed that crowdworkers have consistently lower agreement even on annotating question grammaticality, centrality to topic, and the source of information for the answers. What is worse, neither experts nor crowdworkers were particularly successful with annotating ``types of information needed to answer this question''. \final{But if we do not develop ways to map questions to reasoning steps and target information, we cannot assess whether the systems do what they are supposed to do \cite{ChoudhuryRogersEtAl_2022_Machine_Reading_Fast_and_Slow_When_Do_Models_Understand_Language}.}

The dimension of our taxonomy (\cref{sec:reasoning-taxonomy}) that has received the least attention so far seems to be the logical types of inference. Perhaps not coincidentally, this is the most abstract dimension requiring the most specialist knowledge. However, the criterion of the strength of support for the hypothesis is extremely useful: to be able to trust NLP systems in the real world, we would like to know how they handle reasoning with imperfect information. %
Let us note that at least some inference types map to the question types familiar in QA/RC literature, and could be bootstrapped: 

\begin{itemize*}
\item The questions that involve only interpreting and manipulating (non-ambiguous) linguistic or numerical input fall under deductive reasoning, because the reader is assumed to have a set of extra premises (definitions for words and mathematical operations) shared with the question author.  
\item The questions about the future state of the world, commonsense questions necessarily have a weaker link between the premise and conclusion, and could be categorized as inductive. 
\end{itemize*}

Other question types could target inductive or deductive reasoning, depending on how strong is the evidence provided in the premise: e.g. temporal questions are deductive if the event order strictly follows from the narrative, and inductive if there are uncertainties filled on the basis of script knowledge.

\subsection{What datasets have not been created?}
\label{sec:missing-data}

Notwithstanding all of the numerous datasets in the recent years, the space of unexplored possibilities remains large. Defining what datasets need to be created is itself a part of the progress towards machine RC, and any such definitions will necessarily improve as we make such progress. At this point we would name the following salient directions.

\textbf{Linguistic features of questions and/or the contexts that they target}. The current pre-trained language models do not acquire all linguistic knowledge equally well or equally fast: e.g. RoBERTa~\cite{LiuOttEtAl_2019_RoBERTa_Robustly_Optimized_BERT_Pretraining_Approach} learns the English irregular verb forms already with 100M tokens of pre-training, but struggles with (generally more rare) syntactic island effects even after 1B pre-training~\cite{ZhangWarstadtEtAl_2020_When_Do_You_Need_Billions_of_Words_of_Pretraining_Data}. Presumably knowledge that is less easily acquired in pre-training will also be less available to the model in fine-tuning. There are a few datasets that focus on questions requiring a specific aspect of linguistic reasoning, but there are many untapped dimensions.  How well do our models cope with questions that a human would answer using their knowledge of e.g. scope resolution, quantifiers, or knowledge of verb aspect?

\textbf{Pragmatic properties of questions.} While deixis (contextual references to people, time and place) clearly plays an important role in multimodal and conversational QA resources, there does not seem to be much work focused on that specifically (although many resources cited in \cref{sec:discourse} contain such examples). Another extremely important direction is factuality: there is already much research on fact-checking \cite[][]{hassan2017toward,thorne-etal-2018-fever,augenstein-etal-2019-multifc,atanasova-etal-2020-generating-fact}, but beyond that, it is also important to examine questions for presuppositions \final{and various kinds of ambiguity ambiguity~\cite{MinMichaelEtAl_2020_AmbigQA_Answering_Ambiguous_Open-domain_Questions,ZhangChoi_2021_SituatedQA_Incorporating_Extra-Linguistic_Contexts_into_QA,SunCohenEtAl_2022_ConditionalQA_Complex_Reading_Comprehension_Dataset_with_Conditional_Answers,WangJatowtEtAl_2022_ArchivalQA_Large-scale_Benchmark_Dataset_for_Open_Domain_Question_Answering_over_Historical_News_Collections}}. \final{Another important direction is testing factual correctness of answers coming from pre-training data of large language models \cite{LinHiltonEtAl_2022_TruthfulQA_Measuring_How_Models_Mimic_Human_Falsehoods}.}

\textbf{QA for the social good.} A very important dimension for practical utility of QA/RC data is their domain (\cref{sec:domains}): since domain adaptation is generally very far from being solved, the lack of resources in a given domain may mean that QA systems for that domain are completely unavailable. There are many domains that have not received much attention because they are not backed by commercial interests, and are not explored by academics because there is no ``wild'' data like StackExchange questions that could back it up. %
For instance, QA data that could be used to train FAQ chatbots for education and nonprofit sectors could make a lot of difference for low-resource communities, but is currently notably absent. The same argument extends to data for low-resource languages (see \cref{sec:languages-mono})%

\textbf{Documented limitations.} %
If the training data has statistically conspicuous ``shortcuts'', we have no reason to expect neural nets to not pick up on them~\cite{Linzen_2020_How_Can_We_Accelerate_Progress_Towards_Human-like_Linguistic_Generalization}, and the biases due to e.g. annotation artifacts are actively harmful (since they could prevent the system from generalizing as expected). There are many proposals to improve the data collection methodology (which deserves a separate survey), but it seems that fundamentally we may never be able to guarantee absence of spurious patterns in naturally-occurring data -- and it gets harder as dataset size grows~\cite{GardnerMerrillEtAl_2021_Competency_Problems_On_Finding_and_Removing_Artifacts_in_Language_Data}. The field of AI ethics has drawn much attention to undesirable social biases and called for documenting the speaker demographics \cite{BenderFriedman_2018_Data_Statements_for_Natural_Language_Processing_Toward_Mitigating_System_Bias_and_Enabling_Better_Science,GebruMorgensternEtAl_2020_Datasheets_for_Datasets}. Perhaps we could develop a similar practice for documenting spurious patterns in the data we use, which could then be used for model certification \cite{MitchellWuEtAl_2019_Model_Cards_for_Model_Reporting}. New datasets will be more useful if they come with documented limitations, rather than with the impression that there are none.

\section{Conclusion}

The number of QA/RC datasets produced by the NLP community is large and growing rapidly.  We have presented the most extensive survey of the field to date, identifying the key dimensions along which the current datasets vary. These dimensions provide a conceptual framework for evaluating current and future resources in terms of their format, domain, and target reasoning skills. We have categorized over two hundred datasets while highlighting the gaps in the current literature, and we hope that this survey would be useful both for the NLP practitioners looking for data, and for those seeking to push the boundaries of QA/RC research.

\anna{While quantitative analysis of the possible types of the available QA/RC resources and the way those types are blended was not in scope of this survey, the proposed dimensions of description open up the avenue for such analysis in the future work. In particular, such analysis in diachronic perspective would enable meta-research of the evolution of thinking about machine reading in the NLP community: from the earliest information retrieval benchmarks to the latest resources geared towards few-shot learning. It would also open up the avenue for more sophisticated search interfaces in community data-sharing hubs such as HuggingFace Datasets\footnote{\url{https://huggingface.co/datasets/}}.}

\bibliographystyle{ACM-Reference-Format}
\bibliography{sample-base}

\newpage

\end{document}